\documentclass[10pt,twocolumn]{article}

\usepackage[T1]{fontenc}
\usepackage[utf8]{inputenc}
\usepackage{lmodern}
\usepackage[margin=0.75in,columnsep=0.3in]{geometry}
\usepackage{microtype}
\usepackage{amsmath,amssymb}
\usepackage{graphicx}
\usepackage{booktabs}
\usepackage{multirow}
\usepackage{array}
\usepackage{xcolor}
\usepackage{enumitem}
\usepackage{caption}
\usepackage[numbers,sort&compress]{natbib}
\usepackage{tikz}
\usepackage{pgfplots}
\pgfplotsset{compat=1.18}
\usepgfplotslibrary{statistics}
\usetikzlibrary{arrows.meta,positioning,fit,backgrounds,calc,shapes.symbols,decorations.pathreplacing}
\usepackage[most]{tcolorbox}
\usepackage{xurl}
\usepackage[colorlinks=true,linkcolor=blue!50!black,citecolor=blue!50!black,urlcolor=blue!50!black]{hyperref}

\setlist{itemsep=2pt,topsep=4pt}

\newcommand{\pfdelta}{PF$\Delta$}
\newcommand{\pmci}[1]{\,{\scriptsize$\pm$#1}}

\definecolor{ga1}{HTML}{2F5D8A}
\definecolor{ga2}{HTML}{C0504D}
\definecolor{ga3}{HTML}{9BBB59}
\definecolor{ga4}{HTML}{8064A2}
\definecolor{ga5}{HTML}{F79646}
\definecolor{gagrey}{HTML}{7F7F7F}
\definecolor{exonto}{HTML}{2F5D8A}
\definecolor{exstudio}{HTML}{B35F1F}
\definecolor{vzours}{HTML}{2A78D6}
\definecolor{vzcode}{HTML}{EB6834}
\definecolor{vzgpt}{HTML}{1BAF7A}
\definecolor{vzbase}{HTML}{8A8A8A}
\definecolor{vzagent}{HTML}{7A5AC8}
\pgfplotsset{gaplot/.style={axis lines*=left, tick label style={font=\footnotesize},
  label style={font=\footnotesize}, major grid style={black!8}, axis line style={black!45},
  tick style={black!45}}}

\tcbset{exstyle/.style={enhanced, breakable, boxrule=0.6pt, arc=2pt, left=7pt, right=7pt,
  top=5pt, bottom=5pt, toptitle=3pt, bottomtitle=3pt, fonttitle=\sffamily\bfseries\small,
  fontupper=\small, coltitle=white, before skip=10pt, after skip=10pt}}
\newtcolorbox[auto counter]{example}[3][]{exstyle, colframe=#2, colbacktitle=#2,
  colback=#2!3, title={Example~\thetcbcounter\quad #3}, #1}
\newcommand{\exfield}[1]{\par\smallskip\noindent{\sffamily\bfseries\scriptsize
  \textcolor{black!60}{\MakeUppercase{#1}}}\par\nobreak\smallskip\noindent\ignorespaces}
\newtcolorbox{takeaway}[1]{enhanced, boxrule=0.6pt, arc=3pt, colframe=#1!75, colback=#1!5,
  left=6pt, right=6pt, top=4pt, bottom=4pt, before skip=4pt, after skip=10pt, fontupper=\small,
  before upper={\textbf{Takeaway:}\ }}
\newcommand{\exquote}[1]{\begin{tcolorbox}[enhanced, boxrule=0pt, frame hidden, arc=0pt,
  borderline west={1.5pt}{0pt}{black!25}, colback=black!3, left=6pt, right=4pt, top=3pt,
  bottom=3pt, before skip=2pt, after skip=2pt, fontupper=\small\sloppy]#1\end{tcolorbox}}

\tikzset{person/.pic={\fill (0,0.13) circle (0.075); \fill (-0.12,-0.1) arc[start angle=180,
  end angle=0, radius=0.12] -- cycle;}}

\newcommand{\gamark}{\includegraphics[height=0.72em]{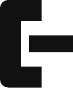}}

\title{\includegraphics[height=0.75cm]{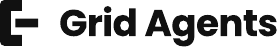}\\[18pt]
\textbf{An Agentic Framework for Automated Power System Studies and Grid Connection Assessment}\\[6pt]
\large Technical Report}
\author{MaoXun Huang \qquad Kshitij Kapoor \qquad Eli Louise\\[4pt]
\normalsize Grid Agents\\
\normalsize\texttt{\{mao,\,kaps,\,eli\}@grid-agents.com}}
\date{} %


\begin{document}

\twocolumn[
\begin{@twocolumnfalse}
\maketitle

\begin{abstract}
Connecting new generation and demand to the grid depends on two scarce skills: checking
applications against a large, changing body of regulation, and running the network studies that
decide where and how a project connects. We present \emph{Grid Agents}, an AI system that brings
both together, and the methodology we use to evaluate it. The \emph{Grid Ontology} is its
knowledge system. It represents the regulation, standards and engineering knowledge that grid
work depends on, makes it searchable, and evolves as the agents learn from use. Every answer drawn
from it rests on evidence the agents cite. \emph{Grid Studio} is an agent system specialised for
power-system engineering. It works in its own engineering environment with domain tools and a
study framework, draws on the Grid Ontology, and relies on established solvers for every number
it reports. In both, engineers make the decisions. We bring eight published power-system
benchmark families, previously scattered across separate papers, into one suite of 519 tasks, and
we verify not only each answer but the work behind it, by replaying the agent's tool calls. The
Grid Ontology cites the right regulatory evidence more often than a general coding agent or
retrieval pipelines, leads on LegalBench-RAG and ranks first on the public leaderboards of the
Enterprise RAG Challenge and ViDoRe V3. Grid Studio scores highest on the 519 power-system tasks
and is the fastest of all the systems we compare, both by a statistically significant
margin. With an optimiser in the loop,
its reinforcement packages cost 33--67\% less on average than those the agent designs alone.
\end{abstract}
{\vspace{0.6em}
\begin{center}
\pgfdeclareplotmark{fstar}{%
  \pgfpathmoveto{\pgfpointpolar{90}{\pgfplotmarksize}}%
  \foreach \a in {126,198,270,342,54} {%
    \pgfpathlineto{\pgfpointpolar{\a}{0.42\pgfplotmarksize}}%
    \pgfpathlineto{\pgfpointpolar{\a+36}{\pgfplotmarksize}}}%
  \pgfpathclose\pgfusepathqfillstroke}
\begin{tikzpicture}[lead/.style={black!35, line width=0.35pt}, lab/.style={font=\scriptsize, inner sep=1.2pt}]
\begin{axis}[gaplot, set layers, width=\textwidth, height=5.8cm, xmode=log, xmin=8, xmax=300,
  ymin=0, ymax=110, grid=major, xtick={10,20,50,100,200}, xticklabels={10,20,50,100,200},
  minor xtick={9,30,40,60,70,80,90,300},
  xlabel={Median agent work time per task (s, log scale)}, ylabel={End-to-end score $E$ (\%)},
  ytick={0,25,50,75,100}, clip=false, legend columns=-1, legend cell align=left,
  legend style={at={(0.5,1.03)}, anchor=south, draw=none, fill=none, font=\scriptsize,
    /tikz/every even column/.append style={column sep=10pt}}]
\addlegendimage{only marks, mark=fstar, mark size=4.6pt, vzours, draw=white, line width=0.4pt}
\addlegendentry{Grid Studio}
\addlegendimage{only marks, mark=diamond*, mark size=3.2pt, vzcode, draw=white, line width=0.4pt}
\addlegendentry{Coding agent}
\addlegendimage{only marks, mark=*, mark size=2.2pt, vzbase!60, draw=white, line width=0.4pt}
\addlegendentry{Base model}
\pgfonlayer{axis background}
\fill[vzours!8] (axis cs:13.89,0) rectangle (axis cs:300,99.90);
\endpgfonlayer
\node[lab, anchor=north east, font=\scriptsize\itshape, text=vzours!70!black] at (axis cs:292,96)
  {Slower than Grid Studio and scoring lower};
\draw[-{Stealth[length=6pt, width=6pt]}, black!35, line width=1.4pt] (axis cs:11.4,42) -- (axis cs:8.9,72)
  node[midway, sloped, above=1pt, font=\scriptsize\itshape, text=black!55] {Better};
\addplot[forget plot, only marks, mark=*, mark size=2.2pt, vzbase!55, draw=white, line width=0.5pt] coordinates {(70.15,32.83) (36.39,28.27) (19.74,13.67) (21.81,21.42) (30.24,20.86) (58.90,20.16) (99.96,18.43) (23.15,17.26) (28.39,17.35)};
\addplot[forget plot, only marks, mark=*, mark size=2.8pt, vzbase, draw=white, line width=0.6pt] coordinates {(162.54,28.29)};
\addplot[forget plot, only marks, mark=*, mark size=2.8pt, vzgpt, draw=white, line width=0.6pt] coordinates {(58.16,40.84)};
\addplot[forget plot, only marks, mark=diamond*, mark size=3.6pt, vzcode, draw=white, line width=0.6pt] coordinates {(21.60,99.36) (21.98,85.37) (16.81,99.43)};
\addplot[forget plot, only marks, mark=fstar, mark size=5.4pt, vzours, draw=white, line width=0.6pt] coordinates {(13.89,99.90)};
\node[lab, anchor=east, font=\scriptsize\bfseries] at (axis cs:12.50,100.5) {Grid Studio};
\node[lab, anchor=west] at (axis cs:23.11,99.4) {Codex CLI};
\node[lab, anchor=west] at (axis cs:23.51,85.4) {Claude Code};
\draw[black!35, line width=0.35pt] (axis cs:16.81,96.6) -- (axis cs:16.81,81);
\node[lab, anchor=north] at (axis cs:16.81,81) {Power-Agent};
\node[lab, anchor=south] at (axis cs:58.16,44.0) {GPT-6.1 sol};
\node[lab, anchor=west] at (axis cs:75.76,32.8) {GPT-5.6 sol};
\node[lab, anchor=south] at (axis cs:162.54,31.5) {Fable 5.1};
\node[lab, anchor=south] at (axis cs:36.39,31.1) {Sonnet 5.5};
\node[lab, anchor=east] at (axis cs:18.16,13.7) {Nova Pro};
\node[lab, anchor=south] at (axis cs:21.81,24.6) {GPT-OSS-120B};
\node[lab, anchor=west] at (axis cs:32.36,21.6) {Kimi K2.5};
\node[lab, anchor=north] at (axis cs:58.90,17.2) {DeepSeek V3.2};
\node[lab, anchor=west] at (axis cs:106.96,18.4) {Qwen3-Coder-Next};
\node[lab, anchor=north west] at (axis cs:24.65,11.2) {Mistral Large 3};
\node[lab, anchor=west] at (axis cs:30.38,15.6) {GLM-4.7};
\draw[lead] (axis cs:23.61,15.7) -- (axis cs:24.65,11.2);
\end{axis}
\end{tikzpicture}

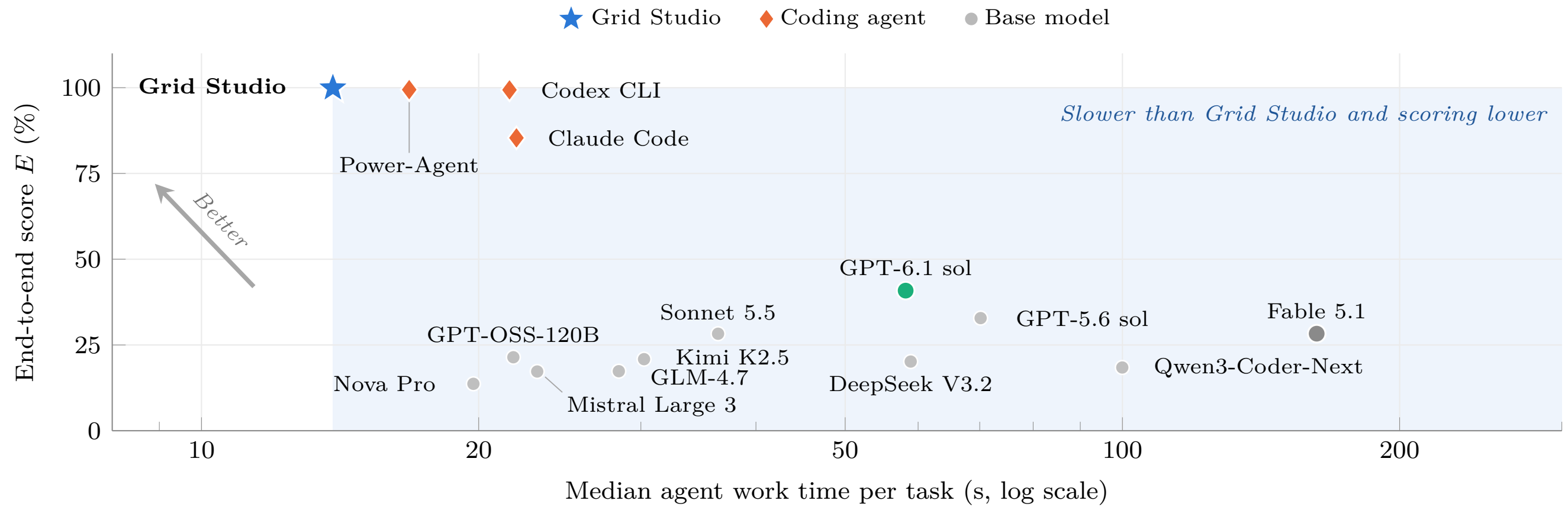
\captionof{figure}{\textbf{End-to-end score against agent work time per task.} Each point is one
system on the same 519 tasks. Every other system is slower than Grid Studio and scores lower, and
both differences are statistically significant (Table~\ref{tab:studio-main}).}
\label{fig:frontier}
\end{center}}
\vspace{1.4em}
\end{@twocolumnfalse}
]

\section{Introduction}
\label{sec:intro}

Grid connection, more than generation, now limits the energy transition. At least 3,000\,GW
of renewable projects were waiting in connection queues worldwide in 2023~\citep{iea2023grids}.
In the United States alone, about 2,600\,GW of generation and storage was seeking interconnection at
the end of that year~\citep{rand2024queuedup}. Both the United States and Great Britain have
reformed their connection processes in response~\citep{ferc2023order2023,ofgem2025connections}.

Each connection request needs two kinds of expert work. \emph{Regulatory} work checks the
application against legislation, grid codes and standards that run to thousands of pages and
change as reform proceeds. \emph{Engineering} work decides where the project can connect and
what reinforcement it needs, by running power-flow, contingency (N-1) and short-circuit studies
on a network model~\citep{kundur1994,wood2013}. Both are slow and both are short of people.

Agents built on large language models (LLMs) plan, call tools and act on the
results~\citep{yao2023react,schick2023toolformer,wang2024agentsurvey}, which suits both tasks.
They also fail in three ways that matter here. They write fluent text that no source
supports~\citep{ji2023hallucination}. They cannot be trusted with the arithmetic of large network
equations. Finally, a correct answer does not show that the work behind it was correct. The same
number can come from the right study, from memory or from a study of the wrong grid state, and
most benchmarks grade only the number. Yet a connection decision commits costly network works and
can be challenged by the applicant or the regulator, so the engineer must be able to show which
study ran on which network.

\begin{figure*}[t]
\centering
\makebox[\textwidth][c]{\begin{tikzpicture}[
  font=\sffamily\small,
  >={Stealth[length=2mm]},
  bandlabel/.style={font=\sffamily\scriptsize\bfseries, text=black!55, align=right, anchor=east},
  card/.style={draw=#1!75!black, fill=#1!13, rounded corners=3pt, align=center,
    line width=0.6pt, inner sep=4pt, execute at begin node={\hyphenpenalty=10000}},
  t/.style={font=\sffamily\small\bfseries},
  arr/.style={->, line width=0.7pt, black!55},
  darr/.style={<->, line width=0.7pt, black!55},
  lbl/.style={font=\sffamily\scriptsize, text=black!65, fill=white, inner sep=1.2pt, rounded corners=1pt}
]
\foreach \y/\h/\name in {0/0.75/PEOPLE, -2.3/0.95/AGENTS, -5.0/1.15/{KNOWLEDGE\\[-1pt]\& TOOLS}} {
  \fill[black!4, rounded corners=4pt] (2.0,\y+\h) rectangle (16.3,\y-\h);
  \node[bandlabel] at (1.82,\y) {\name};
}
\node[card=ga3, minimum width=3.2cm, minimum height=0.95cm] (dev) at (3.9,0)
  {{\bfseries Developer}\\[-1pt]{\scriptsize submits an application}};
\node[card=ga3, minimum width=3.2cm, minimum height=0.95cm] (rev) at (7.6,0)
  {{\bfseries Reviewer}\\[-1pt]{\scriptsize grid operator \textbullet\ decides}};
\node[card=ga3, minimum width=3.6cm, minimum height=0.95cm] (eng) at (13.9,0)
  {{\bfseries Planning engineer}\\[-1pt]{\scriptsize runs studies \textbullet\ decides}};
\node[card=ga1, minimum width=5.6cm, minimum height=1.25cm] (onto) at (5.6,-2.3)
  {{\gamark\hspace{0.4em}\bfseries Grid Ontology agent}\\[1pt]{\scriptsize answers and reviews with citations}};
\node[card=ga1, minimum width=4.6cm, minimum height=1.25cm] (studio) at (13.9,-2.3)
  {{\gamark\hspace{0.4em}\bfseries Grid Studio agent}\\[1pt]{\scriptsize plans and runs studies}};
\draw[ga1!70!black, dashed, line width=0.6pt, rounded corners=4pt] (2.2,-3.98) rectangle (12.35,-6.02);
\node[font=\sffamily\scriptsize\bfseries, text=ga1!80!black, anchor=north west] at (2.32,-4.02) {Grid Ontology};
\node[card=gagrey, text width=4.2cm, minimum height=1.15cm] (corpus) at (4.8,-5.3)
  {{\footnotesize\bfseries Regulation corpus}\\[-1pt]{\scriptsize legislation, codes and standards}};
\node[card=ga4, text width=4.2cm, minimum height=1.15cm] (mem) at (9.75,-5.3)
  {{\footnotesize\bfseries Memory}\\[-1pt]{\scriptsize facts, skills, experiences}};
\node[card=ga5, text width=3.45cm, minimum height=1.55cm] (solv) at (14.45,-5.0)
  {{\footnotesize\bfseries Power-system solvers}\\[1pt]{\scriptsize power flow \textbullet\ N-1 \textbullet\ OPF}\\{\scriptsize short circuit \textbullet\ dynamics}\\{\scriptsize time series}};
\draw[arr] (dev.south) -- node[lbl]{application} (dev.south |- onto.north);
\draw[darr] (rev.south) -- node[lbl]{findings \textbullet\ sign-off} (rev.south |- onto.north);
\draw[darr] (eng.south) -- node[lbl]{consults \textbullet\ decisions} (eng.south |- studio.north);
\draw[darr] (onto.south) -- node[lbl]{retrieve \textbullet\ cite} (onto.south |- 0,-3.98);
\draw[darr] (studio.south) -- node[lbl]{tool calls \textbullet\ results} (studio.south |- solv.north);
\draw[darr] (mem.north) |- node[lbl, pos=0.25]{recall \textbullet\ learn (approved)} (studio.west);
\end{tikzpicture}}
\caption{\textbf{High-level system design of Grid Agents.} People sit above two agents. The Grid
Ontology agent answers and reviews with citations, and Grid Studio runs engineering studies through established
solvers. Both draw on the Grid Ontology's knowledge (the regulation corpus and a memory that learns
from use under approval).}
\label{fig:overview}
\end{figure*}
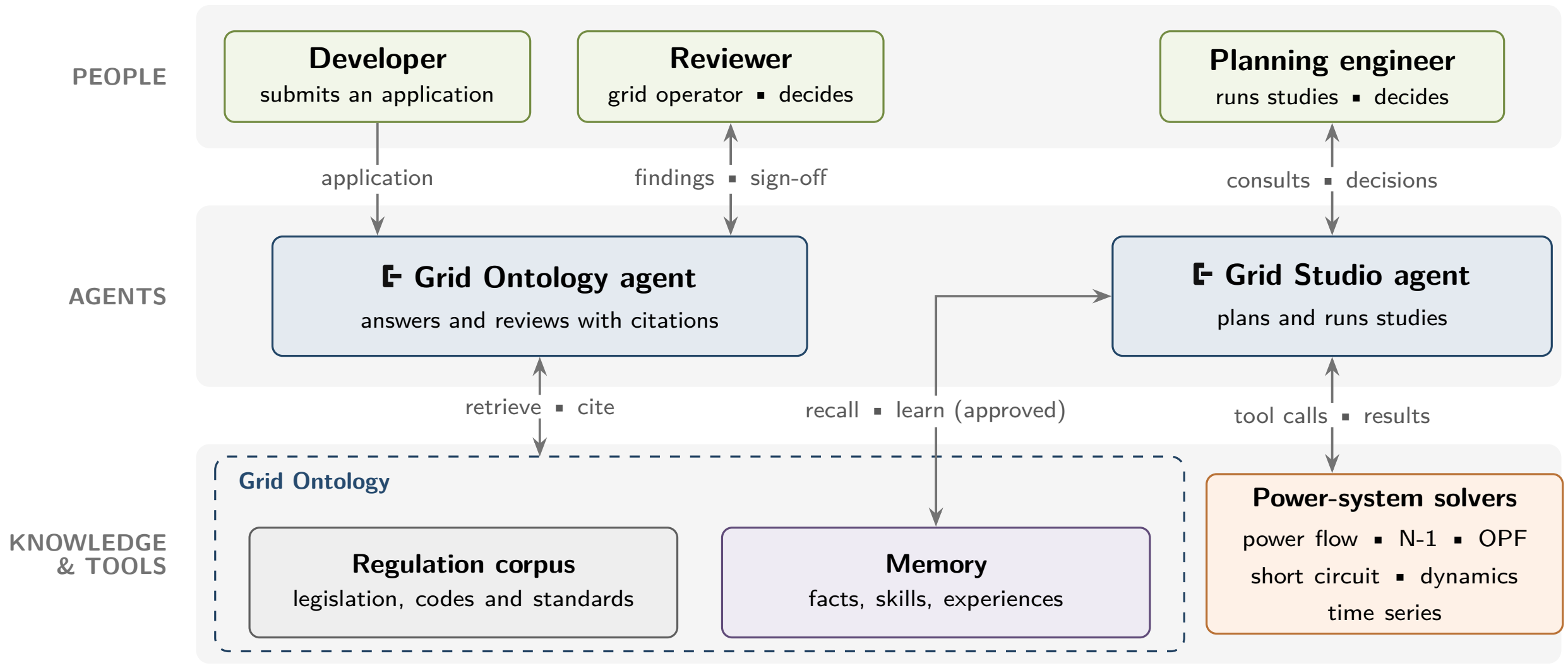

We present \emph{Grid Agents}, a system designed around these failure modes. It has two parts (Figure~\ref{fig:overview}):
\begin{itemize}
  \item The \textbf{Grid Ontology} is the knowledge system of Grid Agents. It represents the
  regulation, standards and engineering knowledge that grid work depends on, makes it searchable
  by the agents, and evolves as they learn from use, with people approving what is
  shared.\footnote{We use \emph{ontology} for an organised body of domain knowledge, not a
  formal specification of a conceptualisation~\citep{gruber1993ontology}.} Agents answer
  questions and review applications from it, and every answer rests on evidence they cite.
  \item \textbf{Grid Studio} is an agent system specialised for power-system engineering. It works
  in its own engineering environment, with domain tools and a framework that carries a study from
  the choice of connection point to the final report, and it is connected to the Grid Ontology for
  the knowledge its studies draw on. Established solvers compute every number it reports.
\end{itemize}

\paragraph{Contributions.}
\begin{enumerate}
  \item \textbf{A design} for agents in regulated engineering work, built on three principles,
  with a memory that learns from use under human approval (Section~\ref{sec:system}).
  \item \textbf{A power-system benchmark suite} of 519 tasks from eight published benchmark
  families, on networks of 4 to 9,241 buses, graded the same way on one scale of depth
  (Section~\ref{sec:benchmarks}).
  \item \textbf{Results and findings} against coding agents, retrieval pipelines, leaderboard
  entries and frontier models, with each agent's work verified against the study it ran
  (Section~\ref{sec:results}). The Grid Ontology cites the right regulatory evidence more often
  than any other system. Grid Studio leads the best coding agent in both score and agent work time,
  by a statistically significant margin, and its lead is largest on the biggest networks.
  Optimisation in the loop makes reinforcement cheaper.
\end{enumerate}

\section{Background and Related Work}
\label{sec:related}

\paragraph{Retrieval and citation.} Retrieval-augmented generation conditions a model on passages fetched for the
question~\citep{lewis2020rag}. In its common form a pipeline searches once and the model answers
from what comes back. The search may be lexical or dense, often followed by a
reranker~\citep{robertson2009bm25,karpukhin2020dpr,nogueira2019passage}. It may also run over a
graph built from the corpus~\citep{edge2024graphrag} or follow a document's own
hierarchy~\citep{pageindex}. Longer context windows do not remove the need to retrieve
well~\citep{liu2024lostmiddle}. Agentic approaches instead let the model decide what to retrieve
and judge what it found~\citep{asai2024selfrag}, or explore the corpus from a programming
environment~\citep{zhang2025rlm}. Because a reviewer must be able to check every source, the
citations matter as much as the answer. Work on attributed generation measures whether citations
support the claims~\citep{gao2023alce}, and model judges score faithfulness and answer
quality~\citep{es2024ragas,zheng2023judge}. LegalBench-RAG~\citep{pipitone2024legalbenchrag},
built on CUAD~\citep{hendrycks2021cuad} and related to LegalBench~\citep{guha2023legalbench},
scores retrieval on exact character spans, which separates finding the right document from
finding the right clause.

\paragraph{Tool-using agents and their evaluation.} Engineering work needs an agent that acts as well as reads. ReAct interleaves reasoning and
action~\citep{yao2023react}, Toolformer learns when to call a tool~\citep{schick2023toolformer},
Reflexion adds self-critique~\citep{shinn2023reflexion}, AutoGen composes several
agents~\citep{wu2023autogen} and MANTA lets a multi-agent system adapt how its agents collaborate
while it runs~\citep{huang2026manta}. As agents began to act, their evaluation moved from judging text to
checking outcomes. SWE-bench runs a repository's tests~\citep{jimenez2024swebench}, AgentBench
spans many environments~\citep{liu2024agentbench} and $\tau$-bench compares the final database
state with a goal state~\citep{yao2024taubench}. Our strict pass brings this state-based grading
to engineering studies by checking the grid state that each study ran on.

\paragraph{Agent memory.} An agent that works beside the same people for months should also keep what it learns. CoALA
organises a language agent's memory into working, semantic, episodic and procedural
stores~\citep{sumers2024coala}, after the distinction drawn in human
memory~\citep{tulving1972episodic}. MemGPT pages information between the context and external
storage~\citep{packer2023memgpt}. Generative Agents and Reflexion turn experience into reflections
that guide later behaviour~\citep{park2023generative,shinn2023reflexion}, ExpeL distils insights
from past trajectories~\citep{zhao2024expel} and Voyager grows a library of
skills~\citep{wang2024voyager}. These systems learn on their own. In regulated work, one
engineer's session must not silently change a colleague's agent, so the memory we describe in
Section~\ref{sec:ontology} adds human approval.

\paragraph{Power-system studies.} In engineering work, the actions an agent takes are calls to power-system analyses whose methods
are long established. A power flow solves the AC network equations for voltages and flows at one
operating point~\citep{wood2013}. Security standards require the system to survive any single
credible outage~\citep{neso_sqss}, so a contingency (N-1) analysis repeats the power flow with
each element out of service. Short-circuit, stability and dynamic studies cover fault levels and
the seconds after a disturbance~\citep{kundur1994}, and optimal power flow (OPF) finds the
least-cost dispatch within network limits. Open-source solvers such as
MATPOWER~\citep{zimmerman2011matpower}, pandapower~\citep{thurner2018pandapower} and
PowerModels.jl~\citep{coffrin2018powermodels} implement these analyses, and several benchmarks
take their answer keys from them. For an engineer, and for an agent, the difficulty lies less in
the computation than in choosing which studies to run on which network state, and in reading the
results against the rules.

\paragraph{Language models for power systems.} Language models have only recently been applied to this work. Surveys map the opportunities and
risks~\citep{huang2023foundation,majumder2024llmenergy,ghafari2025llmreview}. Agent systems
include multi-agent simulation frameworks~\citep{jia2024powersim}, agents for analysis and
operations~\citep{gridmind,xgridagent}, an agent for connection impact
assessment~\citep{shamseldein2026gridmind} and tool servers for power-system
software~\citep{powermcp}. Several are evaluated on task sets of their
own~\citep{gridmind,xgridagent}, which makes them hard to compare. Shared benchmarks have
followed. They cover power-flow questions (PFBench~\citep{pfbench}), agentic studies
(PowerAgentBench~\citep{poweragentbench}, PSAB~\citep{psab}), natural-language study tasks
(PowerCodeBench~\citep{powercodebench}), power flow under load, dispatch and topology variation
(\pfdelta~\citep{pfdelta}, solved with Ipopt~\citep{wachter2006ipopt}) and natural-language OPF
modifications (ProOPF~\citep{proopf}). Older datasets supply independent references, among them
PGLib-OPF~\citep{babaeinejadsarookolaee2019pglib}, OPFData~\citep{lovett2024opfdata},
Grid2Op/L2RPN and RL2Grid~\citep{donnot2020grid2op,marot2020l2rpn,marchesini2025rl2grid} and the
IEEE PES small-signal benchmark systems~\citep{canizares2017benchmark}.

\section{System Overview}
\label{sec:system}

\subsection{Design principles}

\begin{description}
  \item[P1. Evidence before answers.] Every answer and finding is anchored to a passage in a
  source document or in the applicant's submission.
  \item[P2. Computation by established solvers.] The model chooses, sequences and interprets
  studies, and solvers produce every number.
  \item[P3. Human decision authority.] Agents prepare the work and people decide. Agent verdicts sit
  beside human verdicts, decisions need sign-off by named people, and studies pause at decision
  points~\citep{parasuraman2000automation}.
\end{description}

\subsection{The Grid Ontology}
\label{sec:ontology}

\paragraph{Knowledge base.} The Grid Ontology holds the legislation, grid codes, standards
and process documents that an application is checked against. The GB corpus in our experiments
has 27 documents with 6,347 pages, including their figures and tables.

\paragraph{Grounded answers.} An agent answers questions and reviews applications from this
knowledge base. It answers only from evidence it has found and must cite the passages its answer
relies on.

\paragraph{Memory.} Documents state the rules, and memory holds what the agents learn while
applying them. Following the CoALA taxonomy~\citep{sumers2024coala}, it holds three kinds (Table~\ref{tab:memory}). %

\begin{table}[ht]
\centering
\small
\caption{\textbf{Memory carries what the agents learn beyond the documents.} The three kinds held in the Grid Ontology.}
\label{tab:memory}
\begin{tabular}{@{}l >{\raggedright\arraybackslash}p{6.5cm}@{}}
\toprule
Kind & Holds \\
\midrule
Semantic & Facts about the domain, an operator and its projects, and rules for responding \\
Procedural & Skills, which are written methods for a task such as a connection study \\
Episodic & Experiences and the lessons drawn from them \\
\bottomrule
\end{tabular}
\end{table}

\paragraph{Self-evolving memory.} All three kinds of memory change with use. Lessons drawn
from recent work~\citep{park2023generative,zhao2024expel} become experiences, and facts, rules and
skills are added or revised as the work shows what the agents lack. A change that would affect
other people takes effect only after a person approves it. The agents therefore improve with use
while every shared change stays visible and attributable, which is principle P3 applied to
learning.

\subsection{Grid Studio}
\label{sec:studio}

\paragraph{Studies through established solvers.} Grid Studio gives the agent power-system
studies as tools. The agent chooses which studies to run and on which network state, and reads
the results, while established solvers compute every number (P2).

\paragraph{Reinforcement planning.} When a project needs reinforcement, the agent calls an
optimiser that proposes the least-cost package of works, in the spirit of transmission expansion
planning~\citep{latorre2003tep}. Section~\ref{sec:reinforcement} evaluates it.

\section{Experimental Setup}
\label{sec:setup}

\subsection{Benchmarks}
\label{sec:benchmarks}

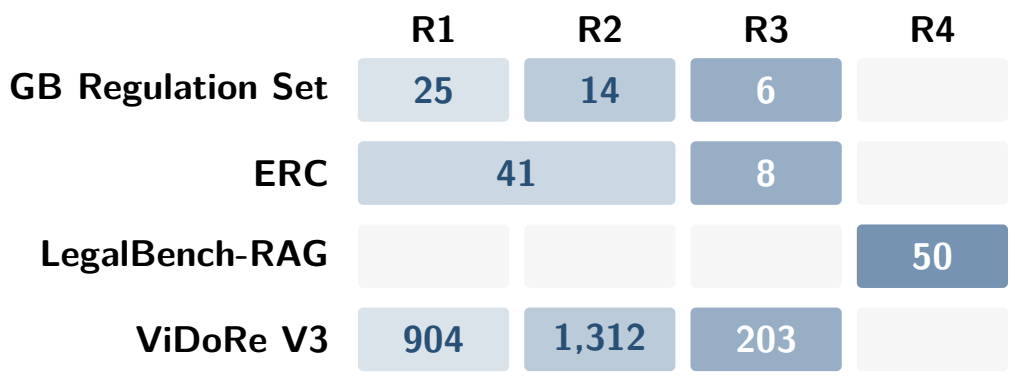
\begin{figure}[t]
\centering
\begin{tikzpicture}[
  font=\sffamily\scriptsize,
  nm/.style={anchor=east, inner sep=1pt, font=\sffamily\scriptsize\bfseries},
  code/.style={anchor=base, font=\sffamily\scriptsize\bfseries},
  num/.style={font=\sffamily\scriptsize\bfseries}
]
\def\xo{2.3}\def\pw{1.1}\def\ph{0.55}\def\cw{0.5}\def\ch{0.21}
\foreach \j in {1,...,4} \node[code] at ({\xo+(\j-0.5)*\pw},0.32) {R\j};
\foreach \k/\name in {1/{GB Regulation Set}, 2/{ERC}, 3/{LegalBench-RAG}, 4/{ViDoRe V3}} {
  \node[nm] at (\xo-0.1,{-(\k-1)*\ph}) {\name};
  \foreach \j in {1,...,4}
    \fill[black!3.5, rounded corners=1pt] ({\xo+(\j-0.5)*\pw-\cw},{-(\k-1)*\ph-\ch}) rectangle
      ({\xo+(\j-0.5)*\pw+\cw},{-(\k-1)*\ph+\ch});
}
\foreach \k/\j/\s/\ink/\n in {1/1/18/exonto!85!black/25, 1/2/32/exonto!85!black/14, 1/3/50/white/6,
    2/3/50/white/8, 3/4/66/white/50, 4/1/18/exonto!85!black/904, 4/2/32/exonto!85!black/{1,312},
    4/3/50/white/203} {
  \fill[exonto!\s, rounded corners=1pt] ({\xo+(\j-0.5)*\pw-\cw},{-(\k-1)*\ph-\ch}) rectangle
    ({\xo+(\j-0.5)*\pw+\cw},{-(\k-1)*\ph+\ch});
  \node[num, text=\ink] at ({\xo+(\j-0.5)*\pw},{-(\k-1)*\ph}) {\n};
}
\fill[exonto!25, rounded corners=1pt] ({\xo+0.5*\pw-\cw},{-\ph-\ch}) rectangle ({\xo+1.5*\pw+\cw},{-\ph+\ch});
\node[num, text=exonto!85!black] at ({\xo+\pw},-\ph) {41};
\end{tikzpicture}
\caption{\textbf{How deep each regulatory benchmark goes.} Rows are the Grid Ontology's
benchmarks and columns the steps of depth. An R1 question is answered by one passage, R2 by a
figure or table, R3 by combining passages and R4 by the exact span of a clause. Each cell counts
the questions at that step, and darker cells are deeper. ERC is the Enterprise RAG Challenge, and
its 51 questions with no answer in the reports are not shown.}
\label{fig:depth-reg}
\end{figure}

Table~\ref{tab:benchmarks} lists the benchmarks and what each tests. We also give every task a
\emph{depth}, how much work the system must do to reach the answer, in the spirit of
PowerCodeBench's difficulty levels~\citep{powercodebench}.
Figures~\ref{fig:depth-reg} and~\ref{fig:depth-power} place each benchmark on this scale, and Table~\ref{tab:depth-items} in
Appendix~\ref{app:examples} gives an example of each step.

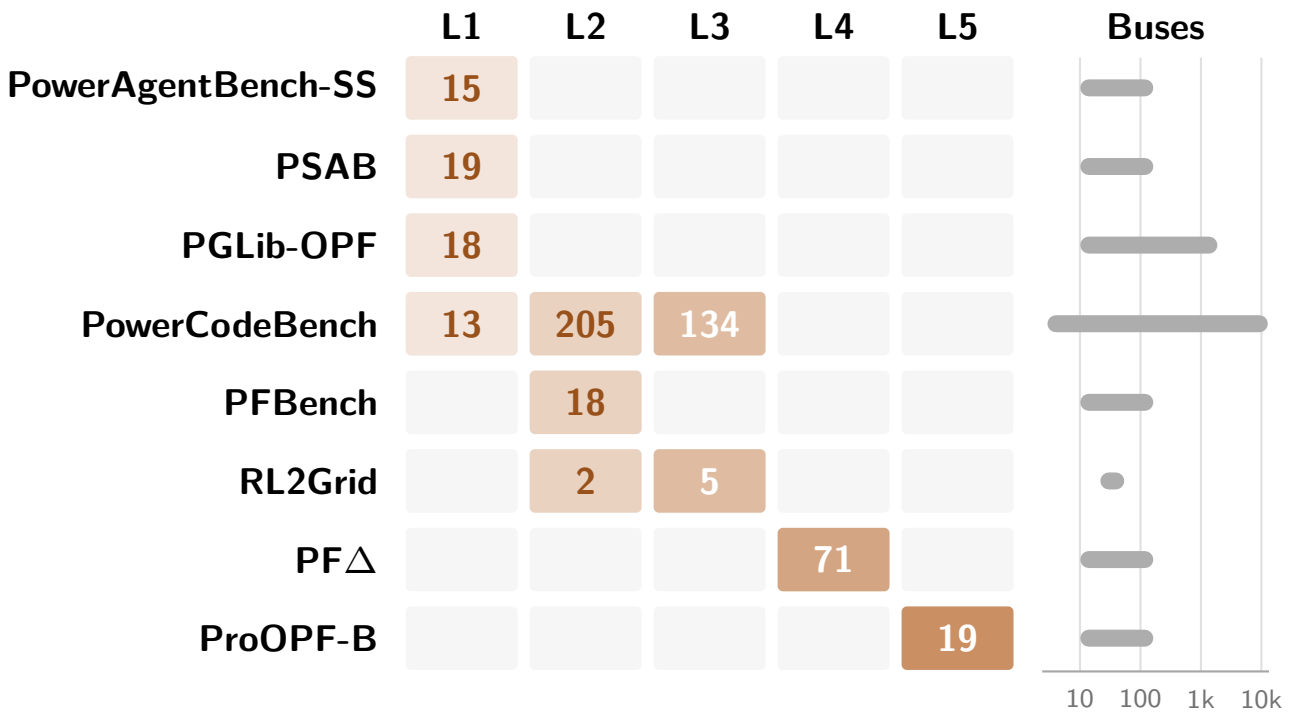
\begin{figure}[t]
\centering
\begin{tikzpicture}[
  font=\sffamily\scriptsize,
  nm/.style={anchor=east, inner sep=1pt, font=\sffamily\scriptsize\bfseries},
  code/.style={anchor=base, font=\sffamily\scriptsize\bfseries},
  num/.style={font=\sffamily\scriptsize\bfseries}
]
\def\xo{2.6}\def\pw{0.82}\def\ph{0.52}\def\cw{0.37}\def\ch{0.21}
\foreach \j in {1,...,5} \node[code] at ({\xo+(\j-0.5)*\pw},0.32) {L\j};
\node[code] at (7.6,0.32) {Buses};
\foreach \d/\lab in {1/10, 2/100, 3/1k, 4/10k} {
  \draw[black!12] ({6.9+0.4*(\d-0.5)},0.2) -- ({6.9+0.4*(\d-0.5)},{-7*\ph-0.22});
  \node[text=black!55, font=\sffamily\tiny, inner sep=0pt] at ({6.9+0.4*(\d-0.5)},{-7*\ph-0.4}) {\lab};
}
\draw[black!35] (6.85,{-7*\ph-0.22}) -- (8.35,{-7*\ph-0.22});
\foreach \k/\name/\lo/\hi in {1/{PowerAgentBench-SS}/14/118, 2/{PSAB}/14/118, 3/{PGLib-OPF}/14/1354,
    4/{PowerCodeBench}/4/9241, 5/{PFBench}/14/118, 6/{RL2Grid}/30/39, 7/{PF$\Delta$}/14/118,
    8/{ProOPF-B}/14/118} {
  \node[nm] at (\xo-0.1,{-(\k-1)*\ph}) {\name};
  \foreach \j in {1,...,5}
    \fill[black!3.5, rounded corners=1pt] ({\xo+(\j-0.5)*\pw-\cw},{-(\k-1)*\ph-\ch}) rectangle
      ({\xo+(\j-0.5)*\pw+\cw},{-(\k-1)*\ph+\ch});
  \draw[black!32, line width=3.2pt, line cap=round]
    ({6.9+0.4*(log10(\lo)-0.5)},{-(\k-1)*\ph}) -- ({6.9+0.4*(log10(\hi)-0.5)},{-(\k-1)*\ph});
}
\foreach \k/\j/\s/\ink/\n in {1/1/16/exstudio!85!black/15, 2/1/16/exstudio!85!black/19,
    3/1/16/exstudio!85!black/18, 4/1/16/exstudio!85!black/13, 4/2/28/exstudio!85!black/205,
    4/3/42/white/134, 5/2/28/exstudio!85!black/18, 6/2/28/exstudio!85!black/2, 6/3/42/white/5,
    7/4/56/white/71, 8/5/70/white/19} {
  \fill[exstudio!\s, rounded corners=1pt] ({\xo+(\j-0.5)*\pw-\cw},{-(\k-1)*\ph-\ch}) rectangle
    ({\xo+(\j-0.5)*\pw+\cw},{-(\k-1)*\ph+\ch});
  \node[num, text=\ink] at ({\xo+(\j-0.5)*\pw},{-(\k-1)*\ph}) {\n};
}
\end{tikzpicture}
\caption{\textbf{How deep each power-system benchmark goes.} Rows are Grid Studio's benchmarks and
columns the steps of depth. An L1 task runs one study on the published network, L2 first applies the
changes it states, L3 must first find what to change, L4 rebuilds a published operating point and L5
turns a written requirement into a changed optimisation problem. Cells count tasks as in
Figure~\ref{fig:depth-reg}, and each bar on the right spans the network sizes a benchmark uses.}
\label{fig:depth-power}
\end{figure}

\begin{table*}[t]
\centering
\small
\caption{\textbf{The benchmarks.} Who developed each one, what it tests and how deep it goes, in the
steps of Figures~\ref{fig:depth-reg} and~\ref{fig:depth-power}. OPF is optimal power flow, the least-cost dispatch within network limits. $^\dagger$Adapted to a network-study interface, so its answer
key comes from a reference solver (Appendix~\ref{app:protocol}).}
\label{tab:benchmarks}
\setlength{\tabcolsep}{5pt}
\begin{tabular}{@{}l l l c r@{}}
\toprule
Benchmark & Developed by & What it tests & Depth & Items \\
\midrule
\emph{Regulatory knowledge} \\
\emph{(Grid Ontology)} \\
GB Regulation Set & Grid Agents (this work) & Questions on GB grid regulation & R1--R3 & 45 \\
Enterprise RAG Challenge~\citep{erc2025} & TimeToAct Austria & Questions on annual reports & R1--R3 & 100 \\
LegalBench-RAG~\citep{pipitone2024legalbenchrag} & ZeroEntropy, UC Berkeley (CUAD) & Exact clause spans in contracts & R4 & 50 \\
ViDoRe V3~\citep{loison2026vidore3} & Illuin Technology, NVIDIA & Pages of visually rich documents & R1--R3 & 2,419 \\
\midrule
\emph{Power-system engineering} \\
\emph{(Grid Studio)} \\
PowerAgentBench-SS$^\dagger$~\citep{poweragentbench} & NTUA, Harvard and others & Single-outage screening (N-1) & L1 & 15 \\
PSAB$^\dagger$~\citep{psab} & Pskov State University & Overload, voltage, outage checks & L1 & 19 \\
PGLib-OPF~\citep{babaeinejadsarookolaee2019pglib} & IEEE PES Task Force & Power flow and OPF & L1 & 18 \\
PowerCodeBench~\citep{powercodebench} & University of Exeter & Study requests in plain English & L1--L3 & 352 \\
PFBench~\citep{pfbench} & Kansas State University & Power flow after load changes & L2 & 18 \\
RL2Grid$^\dagger$~\citep{marchesini2025rl2grid} & MIT, RTE, National Grid ESO & Switching after double outages & L2--L3 & 7 \\
\pfdelta~\citep{pfdelta} & MIT & Rebuilding a published grid state & L4 & 71 \\
ProOPF-B~\citep{proopf} & Peking U., Zhejiang U. & OPF changes from a written brief & L5 & 19 \\
\bottomrule
\end{tabular}
\end{table*}

\paragraph{Regulatory knowledge.} The GB Regulation Set is our own set of 45 questions on GB grid
regulation, asked of the 27 documents of the GB corpus (6,347 pages). An R1 question is answered by one passage, an R2 question by a figure or table, and an
R3 question only by combining passages. We built it evidence first. Passages and figures are
sampled from the corpus, a model writes a question from each, questions the evidence does not
support or that need no evidence are discarded, and the domain owner approves the rest
(Appendix~\ref{app:protocol}). The Enterprise RAG Challenge asks similar questions of 100 company
annual reports (14,454 pages), and about half have no answer in the reports, so the system must also recognise
when there is none. LegalBench-RAG (R4) asks for the exact span of a clause in a long contract, and
we use a fixed sample of 50 questions from its CUAD subset~\citep{hendrycks2021cuad}, whose 462 contracts are plain text without
pages (about 3.9 million words). ViDoRe
V3~\citep{loison2026vidore3} asks 2,419 questions of 19,252 pages of visually rich documents in
English and French, and scores the ranking of the pages a system returns. Its own question labels
give each question a depth (Appendix~\ref{app:protocol}).

\paragraph{Power-system engineering.} Eight benchmark families give 519 tasks on networks of 4 to
9,241 buses. An L1 task runs one study on the network as published, an L2 task first applies stated
changes, and an L3 task must first find what to change, such as the most heavily loaded line. An L4
task rebuilds a published operating point of 20 to 261 values, and an L5 task turns a written
requirement into a changed optimisation problem. None poses a multi-stage study, so
Section~\ref{sec:reinforcement} measures one separately.

\subsection{Evaluation metrics}
\label{sec:metrics}

\paragraph{Regulatory knowledge.} We score the evidence each system \emph{cites}. A cited item
matches a gold anchor if it comes from the same document and overlaps the anchor's span, lies
within one page of it, or is the anchored figure. We report recall, precision, MRR,
nDCG@10~\citep{jarvelin2002ndcg} and hit rate, and a model judge marks each answer against the gold
answer~\citep{zheng2023judge}. LegalBench-RAG and the Enterprise RAG Challenge keep their own
metrics and scoring script, and ViDoRe V3 its board's nDCG@10, averaged over its sets.

\paragraph{Power-system engineering.} A correct number can come from memory, a lucky guess or a
study on the wrong grid state, so we grade the work behind each answer as well as the answer. A
task is a \emph{strict pass} only if its answer is within the benchmark's tolerance and the
system's recorded work shows that the required study ran on the required grid state. With numeric
closeness $\bar c$, answer accuracy $A$ and strict pass rate $P$, each taken over all $n$ tasks, the
end-to-end score is
\begin{equation}
  E \;=\; 0.25\,\bar c \;+\; 0.25\,A \;+\; 0.5\,P .
  \label{eq:e2e}
\end{equation}
A base model, which answers without tools, gets the network as text in its prompt and reports
the solved states its answer rests on, which we check against our own solution.
Appendix~\ref{app:metrics} defines each term, describes how the work of each kind of system is
verified, and gives the reasons for the weights.

\begin{table*}[ht]
\centering
\small
\caption{\textbf{Grid Ontology results on four retrieval benchmarks.} (a) Every system uses the
same model and is scored on the evidence it cites. F1 is shown with its 95\% confidence interval,
and a dash marks a system not run. (b) The top published entries on each public leaderboard, by
the leaderboard's own score (out of 150 for the Enterprise RAG Challenge, and $\times$100 for
ViDoRe V3). Best in \textbf{bold}.}
\label{tab:ontology}
\setlength{\tabcolsep}{4pt}
\begin{tabular*}{\textwidth}{@{\extracolsep{\fill}}l c c c c c c c c@{}}
\toprule
\textit{(a) Same model} & \multicolumn{5}{c}{GB Regulation Set} & \multicolumn{3}{c@{}}{LegalBench-RAG} \\
\cmidrule(lr){2-6}\cmidrule(l){7-9}
System & Recall\,$\uparrow$ & nDCG\,$\uparrow$ & MRR\,$\uparrow$ & Hit rate\,$\uparrow$ & Answer acc.\,$\uparrow$ & Precision\,$\uparrow$ & Recall\,$\uparrow$ & F1\,$\uparrow$ \\
\midrule
Grid Ontology & \textbf{0.967} & \textbf{0.945} & \textbf{0.944} & \textbf{0.978} & \textbf{1.000} & \textbf{0.980} & 0.980 & \textbf{0.980}$_{\pm 0.030}$ \\
Claude Code (file tools) & 0.856 & 0.863 & 0.878 & 0.889 & \textbf{1.000} & 0.900 & 0.900 & 0.900$_{\pm 0.080}$ \\
Vector + reranker~\citep{karpukhin2020dpr,robertson2009bm25,nogueira2019passage} & 0.922 & 0.917 & 0.933 & 0.956 & 0.889 & 0.820 & \textbf{1.000} & 0.869$_{\pm 0.057}$ \\
PageIndex~\citep{pageindex} & 0.400 & 0.389 & 0.378 & 0.400 & 0.444 & 0.762 & 0.920 & 0.799$_{\pm 0.090}$ \\
MS GraphRAG~\citep{edge2024graphrag} & 0.344 & 0.347 & 0.367 & 0.378 & 0.333 & -- & -- & -- \\
Recursive LM~\citep{zhang2025rlm} & -- & -- & -- & -- & -- & 0.860 & 0.860 & 0.860$_{\pm 0.100}$ \\
\midrule
\multicolumn{9}{@{}l}{\textit{(b) Top published entries}} \\[1pt]
\multicolumn{2}{@{}l}{Enterprise RAG Challenge} & Score\,$\uparrow$ & & \multicolumn{4}{l}{ViDoRe V3} & nDCG@10\,$\uparrow$ \\
\cmidrule(r){1-3}\cmidrule(l){5-9}
\multicolumn{2}{@{}l}{Grid Ontology} & \textbf{124.5} & & \multicolumn{4}{l}{Grid Ontology} & \textbf{70.23} \\
\multicolumn{2}{@{}l}{Ilia Ris (1st)} & 123.7 & & \multicolumn{4}{l}{VisAgentRet (1st)} & 70.01 \\
\multicolumn{2}{@{}l}{Emil Shagiev} & 121.6 & & \multicolumn{4}{l}{NeMo Retriever, agentic} & 69.22 \\
\multicolumn{2}{@{}l}{Dmitry Buykin} & 117.5 & & \multicolumn{4}{l}{EVIE-4.5B + JEV-27B-VL} & 68.25 \\
\multicolumn{2}{@{}l}{Sergey Nikonov} & 116.4 & & \multicolumn{4}{l}{jina-embeddings-v4 + zerank-2} & 65.62 \\
\bottomrule
\end{tabular*}
\end{table*}

\subsection{Evaluation method}
\label{sec:method}

\paragraph{Systems compared.} On the regulatory benchmarks we compare the Grid Ontology with a
general coding agent (Claude Code) that has the corpus as files and its own file and shell tools,
with one-shot retrieval pipelines that search once and answer from their top eight results, and on
the Enterprise RAG Challenge and ViDoRe V3 with their published entries. On the power-system benchmarks we compare
Grid Studio with two coding agents as shipped, Claude Code and Codex CLI on GPT-6.1 sol, each with a
shell, Python and the network as a file. We also compare it with
Power-Agent~\citep{powermcp,powerskills}, which installs PowerMCP's pandapower tools and PowerSkills'
playbooks into Claude Code, and with eleven base models, which answer without tools from the
network given as text, among them the frontier models GPT-6.1 sol and Fable 5.1. Claude Code and
Power-Agent run on the same model as our agents, and Power-Agent uses its tools and playbooks as
their authors publish them. We also time each
system's agent work, the time it spends on model turns and tool runs. Appendix~\ref{app:protocol}
gives the evaluation protocol.

\section{Results}
\label{sec:results}

\subsection{Grid Ontology}
\label{sec:results-ontology}

\begin{takeaway}{exonto}
The Grid Ontology cites the right regulatory evidence more often than any other system (0.967
recall). It also leads on LegalBench-RAG (0.980 F1) and ranks first among published entries
on the Enterprise RAG Challenge and ViDoRe V3.
\end{takeaway}

Table~\ref{tab:ontology} gives the results on all four benchmarks. On the GB Regulation Set and
LegalBench-RAG, every system answers with the same model, so the differences come from the evidence
each finds and cites. On the Enterprise RAG Challenge and ViDoRe V3, the Grid Ontology is first on
the public board.

\paragraph{GB Regulation Set.} Both agents answer every question correctly, so the evidence they
cite is what separates them. The Grid Ontology cites the right evidence for more questions than
Claude Code (recall 0.967 against 0.856), and the gap is widest on single-passage questions that
ask for a number (1.000 against 0.625). This matters in review, where a correct answer that cites
the wrong passage still has to be checked by hand. The retrieval pipelines answer fewer questions
correctly. Vector retrieval with a reranker comes closest, while PageIndex and GraphRAG find the
evidence for fewer than half of the questions. Example~\ref{ex:golden} in
Appendix~\ref{app:examples} shows one question.

\paragraph{LegalBench-RAG.} The Grid Ontology reaches an F1 of 0.980, ahead of every baseline. It
cites one contract per question, while the retrieval pipelines cite up to five, so the right
contract is usually among theirs but so are wrong ones. Its one miss is a contract filed twice,
where it cited the signed copy and the answer key names the unsigned draft. At the benchmark's
stricter clause level, its lead over the pipelines is an order of magnitude
(Appendix~\ref{app:legalbench}).

\paragraph{Enterprise RAG Challenge.} The challenge's own script scores both the exact answers and
the pages they cite, out of 150~\citep{erc2025}. The Grid Ontology scores 124.5 against 123.7 for the best published entry, and
it correctly declines 45 of the 51 questions that the reports cannot answer.

\paragraph{ViDoRe V3.} The board scores the first ten pages returned for each English question by
nDCG@10, averaged over eight public sets. The
Grid Ontology scores 70.23 against 70.01 for the best published pipeline and leads on four of the
eight sets (Appendix~\ref{app:vidore}).

\subsection{Grid Studio}
\label{sec:results-studio}

\begin{takeaway}{exstudio}
Grid Studio scores highest, 99.9\% with verified work on 516 of the 519 tasks, and it is the
fastest system, at a median of 13.9\,s of agent work per task. Both leads are statistically
significant. With optimisation in the loop, its reinforcement packages are up to 67\% cheaper.
\end{takeaway}

Table~\ref{tab:studio-main} gives the mean end-to-end score over the eight benchmark families, and
Table~\ref{tab:studio-full} gives each family. Grid Studio scores 99.9\%, with 100\% on seven of the
eight families and 99.2\% on PowerCodeBench, statistically significantly above every other system.
Codex CLI scores 99.4\%, but 19 of its 23 misses are on networks above 2,000 buses, and it takes
up to 55\% longer per task. Power-Agent also scores 99.4\%, and Grid Studio scores higher on 13
tasks and lower on one, with Grid Studio's time 0.75 of Power-Agent's on the same task. Claude Code
scores 85.4\%.

\begin{table}[t]
\centering
\small
\caption{\textbf{End-to-end score and agent work time on the eight power-system benchmark
families (519 tasks).} Mean $E$ (Eq.~\ref{eq:e2e}) and median time per task, $\pm$ the half-width of the
95\% confidence interval. $^\dagger$\,Statistically significantly better than every other system ($p<0.001$,
paired tests on the same tasks, Holm-corrected).}
\label{tab:studio-main}
\setlength{\tabcolsep}{3pt}
\begin{tabular*}{\columnwidth}{@{\extracolsep{\fill}}l r r@{}}
\toprule
System & Mean $E$ (\%)\,$\uparrow$ & Median time (s)\,$\downarrow$ \\
\midrule
Grid Studio & $\mathbf{99.9}$\pmci{0.1}$^\dagger$ & $\mathbf{13.9}$\pmci{0.4}$^\dagger$ \\
Codex CLI & $99.4$\pmci{0.3}\phantom{$^\dagger$} & $21.6$\pmci{0.7}\phantom{$^\dagger$} \\
Claude Code & $85.4$\pmci{3.2}\phantom{$^\dagger$} & $22.0$\pmci{0.9}\phantom{$^\dagger$} \\
Power-Agent & $99.4$\pmci{0.3}\phantom{$^\dagger$} & $16.8$\pmci{0.5}\phantom{$^\dagger$} \\
GPT-6.1 sol, base model & $40.8$\pmci{2.9}\phantom{$^\dagger$} & $58.2$\pmci{10.6}\phantom{$^\dagger$} \\
Fable 5.1, base model & $28.3$\pmci{3.0}\phantom{$^\dagger$} & $162.5$\pmci{39.6}\phantom{$^\dagger$} \\
\bottomrule
\end{tabular*}
\end{table}

\paragraph{Base models.} The best base model, GPT-6.1 sol, answers 250 of the 519 tasks (48\%)
correctly. Given the network as text, the strongest base models solve small power flows by hand,
but none has its work verified on a network of more than 39 buses. None answers more than 30\% of
the \pfdelta{} tasks, where a published operating point must be solved, or more than two of the
seven RL2Grid tasks, where switching actions must be compared, and none scores above 41\% end to
end.

\paragraph{Score against time.} Figure~\ref{fig:frontier} sets each score
against the agent work time. Grid Studio needs a median of 13.9\,s per task, against at most
21.6\,s for Codex CLI and 22.0\,s for Claude Code, so it scores highest in the least time. Reading
the network and writing out the solved states takes the base models a median of 20\,s to almost
3 minutes per task, so every other system is both slower than Grid Studio and lower scoring. More
time does not close the gap: GPT-6.1 sol and Fable 5.1 spend about 1 and 3 minutes per task and
still stay below 41\%.

\paragraph{Domain tools must reach the study.} Power-Agent installs PowerMCP's pandapower tools and
PowerSkills' playbooks into Claude Code, on the same model as Grid Studio, and comes close to Grid
Studio's score. Its PowerMCP tools were called on only 46 of the 519 tasks and settled 16 on their
own. The other 503 needed the agent's own pandapower code, and 473 never called PowerMCP at all,
while a PowerSkills playbook was followed in 409. Its results therefore come mostly from code the
agent writes for each task rather than from the domain tools. Grid Studio's tools reach every study
the benchmarks ask for, which is what makes its work both verifiable and fast.

\paragraph{Real studies are larger.} The benchmark networks have at most 9,241 buses. An operator's
planning model can run to tens of thousands of buses, and a connection study chains many studies,
from screening the options to designing reinforcement. The baselines already fall short at the top
of the benchmarks. Of the 109 tasks on networks above 2,000 buses, Power-Agent passes 106, Codex CLI
90 and Claude Code 74, against 108 for Grid Studio, and the base models pass none. Grid Studio takes its numbers from established solvers rather than from code written
for each task, and its work can be checked at every step, which makes it the better design for real
planning studies.

\begin{table}[t]
\centering
\small
\caption{\textbf{Cost of the reinforcement packages.} Cost in US\$ million, the mean of two runs,
of the package the agent designed alone and with the optimiser.}
\label{tab:reinforcement}
\begin{tabular}{@{}l r r r@{}}
\toprule
Network & Project & \shortstack[r]{Agent\\alone\,$\downarrow$} & \shortstack[r]{With\\optimiser\,$\downarrow$} \\
\midrule
30-bus & 400\,MW & 151.0 & \textbf{81.6} \\
118-bus & 500\,MW & 173.3 & \textbf{71.1} \\
118-bus & 700\,MW & 395.3 & \textbf{166.4} \\
118-bus, three cases & 700\,MW & 481.0 & \textbf{160.4} \\
2,000-bus & 1,100\,MW & 512.7 & \textbf{344.5} \\
\bottomrule
\end{tabular}
\end{table}

\paragraph{Optimisation makes reinforcement cheaper.}
\label{sec:reinforcement}
The benchmarks pose single studies, so we tested the reinforcement stage of a connection study
separately. In five studies on networks of 30, 118 and 2,000 buses, the last the synthetic Texas
grid ACTIVSg2000~\citep{birchfield2017synthetic}, the agent designed the package twice on its own
and twice with the optimiser, using the same model and catalogue of priced reinforcement works
(Table~\ref{tab:reinforcement}). Both approaches cleared every overload the projects caused, as an
independent AC N-1 check of each final network confirms. The optimiser's packages were 33--67\%
cheaper on average.

\section{Conclusion}
\label{sec:conclusion}

Grid Agents is built on three principles. Answers rest on cited evidence, numbers come from
established solvers, and engineers make the decisions. The Grid Ontology cites the right
regulatory evidence more often than a general coding agent or retrieval pipelines, leads on
LegalBench-RAG, and ranks first on the public leaderboards of the Enterprise RAG Challenge and
ViDoRe V3. On 519 tasks from eight published power-system benchmark families, Grid Studio scores
highest and needs the least agent work time, both by a statistically significant margin. With an optimiser
in the loop, its reinforcement packages cost 33--67\% less on average than those the agent designs alone.

\section*{Acknowledgements}

We thank the authors of every benchmark used here for releasing their questions, answer keys and
reference solutions, and TimeToAct Austria, Illuin Technology and NVIDIA for running the public
leaderboards of the Enterprise RAG Challenge and ViDoRe V3. The GB network models come from the
University of Edinburgh Power Systems Test Case Archive, and Manolis Belivanis of the University of
Strathclyde provided the reduced GB model. The synthetic Texas grid comes from the Texas A\&M
Electric Grid Test Case Repository, and the PGLib-OPF cases from the IEEE PES Task Force on
Benchmarks for Validation of Emerging Power System Algorithms.

\bibliographystyle{plainnat}
\bibliography{references}

\clearpage
\appendix
\twocolumn[{%
  \centering
  {\LARGE\bfseries Appendix}\par\vspace{14pt}
  \rule{\textwidth}{0.4pt}\par\vspace{16pt}
}]
\phantomsection\label{app:start}
\section{Example Questions and Tasks}
\label{app:examples}

Table~\ref{tab:depth-items} gives one question per depth level. Examples~\ref{ex:golden}
and~\ref{ex:cuad} are regulatory questions, and Examples~\ref{ex:pfdelta} to~\ref{ex:rl2grid} are
power-system tasks with their strict-pass checks.

\begin{table*}[t]
\centering
\small
\caption{\textbf{One question per depth level (abridged).} Questions in quotation marks are as
the system receives them, and the others are paraphrased. Answer keys are the benchmarks' own.}
\label{tab:depth-items}
\begin{tabular}{@{}>{\raggedright\arraybackslash}p{2.7cm} >{\raggedright\arraybackslash}p{10.6cm} >{\raggedright\arraybackslash}p{3.2cm}@{}}
\toprule
Depth level & Question & Answer key \\
\midrule
R1. Single passage & ``In the RIIO-3 Final Determinations for Electricity Transmission, what are `user progression milestones' as referenced in the CUSC?'' & Conditional and Construction Progression Milestones \\ \addlinespace[3pt]
R2. Figure or table & ``According to the G99 connection process diagram for Types A-D PGMs, how many sequential phases are shown in the flowchart, \dots?'' & Six phases \\ \addlinespace[3pt]
R3. Several passages & ``Under the Grid Code, \dots\ what is the post-fault voltage recovery level (Urec2) at tclear according to Figure ECC.6.3.15.7, and at what operating condition must the equipment be capable of satisfying the Fault Ride Through requirements \dots?'' & 0.85\,pu at 0.14\,s, at rated MW and maximum leading power factor \\ \addlinespace[3pt]
R4. Clause spans & ``Consider the Blockchain Administration and Development Agreement between ARCA U.S. Treasury Fund and ARCA Capital Management, LLC; What is the renewal term for this contract?'' & A 605-character clause \\
\midrule
L1. As given & ``On the PEGASE 9241-bus, run DC power flow. What is the loading of transformer 1652 in percent?'' & 976.22\% \\ \addlinespace[3pt]
L2. Stated changes & On case118, scale the load at bus 7 by P$\times$0.9705 and Q$\times$1.0312 and the load at bus 56 by P$\times$1.0881 and Q$\times$0.9949, run an AC power flow and report the voltage magnitude at bus 79. & 1.009223\,pu \\ \addlinespace[3pt]
L3. Resolve, then modify & ``On the PEGASE 9241-bus, identify the most heavily loaded line under the current operating point and take that line out of service, run DC power flow. What is the loading of transformer 1602 in percent?'' & 61.62\% \\ \addlinespace[3pt]
L4. Operating point & Rebuild the demand at eleven buses, the dispatch of three units, four voltage set-points and one unit outage on the IEEE 14-bus system, then report the voltage at bus 9 (Example~\ref{ex:pfdelta}). & 0.904925\,pu \\ \addlinespace[3pt]
L5. Reformulate & ``Perform AC optimal power flow (ACOPF) on the IEEE 14-bus system (case14). Scale the active power demand at bus 3 by a factor of 1.8. Set the maximum active power output of all generators at bus 2 to 200\,MW. \dots'' (Example~\ref{ex:proopf}) & Objective 11,178.32 \\
\bottomrule
\end{tabular}
\end{table*}

\begin{example}[label=ex:golden]{exonto}{A regulatory question from the GB Regulation Set}
\exfield{Question}
\exquote{According to the TMO4+ Summary Decision Document, what target has the UK Government set for
carbon-free electricity by 2030 as part of its Clean Power 2030 Action Plan?}
\exfield{Answer key}
95\% carbon-free electricity by 2030. Evidence anchor: the TMO4+ Summary Decision Document (Ofgem,
April 2025), page 4, ``\dots\ to achieve 95\% carbon-free electricity by 2030 \dots''
\exfield{Grid Ontology}
Correct, and its first citation is the anchored passage (recall 1.0, reciprocal rank 1).
\exfield{Claude Code with file tools only}
Correct, but it cites nothing, so recall is 0.
\exfield{Why it matters}
Both answers are right, but only one can be checked against its source. Across the set, this is
the recall gap in Table~\ref{tab:ontology}.
\end{example}

\begin{example}[label=ex:cuad]{exonto}{Clause-level evidence on LegalBench-RAG (CUAD)}
\exfield{Question}
\exquote{Consider the Blockchain Administration and Development Agreement between ARCA U.S. Treasury
Fund and ARCA Capital Management, LLC; What is the renewal term for this contract?}
\exfield{Gold clause (605 characters, abridged)}
``This Agreement shall continue in effect for two years from the date hereof, and thereafter shall
continue automatically for successive annual periods, provided that such continuance is
specifically approved at least annually [\dots].''
\exfield{Grid Ontology}
Cites the whole clause (precision, recall and F1 1.00).
\exfield{Claude Code with file tools only}
Gives the same answer but cites only the first 149 characters (precision 1.00, recall 0.25,
F1 0.40), leaving out the annual approval that makes the renewal conditional.
\end{example}

\begin{example}[label=ex:pfdelta]{exstudio}{Rebuilding a published operating point (\pfdelta, IEEE 14-bus, one unit out)}
\exfield{Question (abridged)}
\exquote{Reproduce the operating point below on the base case, exactly as given, then run the study
that answers the question and report the value it solves for. Every figure listed is an input to
build. None of them is the answer.\\[2pt]
\textbullet\ Generating units out of service: gen\_1\\
\textbullet\ Demand by bus, [P MW, Q Mvar]: \{"1": [27.126, 3.916], "2": [40.063, 0.053], \dots\ eleven buses\}\\
\textbullet\ Scheduled active power by unit, MW: \{"gen\_0": 59.0, "gen\_2": 0.0, "gen\_3": 0.0\}\\
\textbullet\ Voltage set-point by bus, pu: \{"0": 1.06, "1": 1.05067, "5": 1.06, "7": 1.06\}\\[2pt]
\dots\ What is the voltage magnitude (per unit) at bus 9?}
\exfield{Answer key}
0.904925\,pu $\pm$ 0.001, from the authors' Ipopt solution.
\exfield{Strict-pass checks}
(1)~The answer is within 0.001\,pu. (2)~The replayed calls show gen\_1 out of service and every
stated demand, dispatch and set-point in force (values the base case already holds may be left as
they are). (3)~An AC power flow was solved after the last change. (4)~Nothing else was changed. A
correct number that fails (2), (3) or (4) is not a strict pass.
\exfield{Result}
Grid Studio passes all 71 \pfdelta{} tasks strictly. Of the two frontier base models,
GPT-6.1 sol answers 29.6\% of them correctly, with its work verified on three 14-bus tasks, and
Fable 5.1 answers 11.3\%.
\end{example}

\begin{example}[label=ex:proopf]{exstudio}{A natural-language OPF modification (ProOPF-B, level 1)}
\exfield{Requirement (the authors' words)}
\exquote{Perform AC optimal power flow (ACOPF) on the IEEE 14-bus system (case14). Scale the active
power demand at bus 3 by a factor of 1.8. Set the maximum active power output of all generators at
bus 2 to 200 MW. Set the OPF violation tolerance (opf.violation) to 1e-6, and write the
corresponding MATPOWER code.}
\exfield{Answer key}
Objective 11,178.32 from the authors' MATPOWER run, at the benchmark's relative tolerance of
$10^{-4}$.
\exfield{Strict-pass checks}
The objective is within tolerance, and the replay shows bus 3's active demand at 1.8 times its base
value (reactive demand unchanged) and the bus-2 unit's maximum at 200\,MW before the OPF is solved.
The solver tolerance is a MATPOWER setting, not part of the network, and is not graded.
\end{example}

\begin{example}[label=ex:rl2grid]{exstudio}{Comparing switching states (RL2Grid adaptation, IEEE 30-bus)}
\exfield{Question (paraphrased)}
\exquote{In case30 at base load, evaluate these simultaneous double-line outages independently:
(L9, L28), (L28, L29) and (L28, L15). Which produces the highest maximum branch loading, and what is
that loading?}
\exfield{Answer key}
The pair (L9, L28), at 154.5\% $\pm$ 1 point, from the reference solver (an adapted family,
Table~\ref{tab:benchmarks}).
\exfield{Strict-pass checks}
Each pair is built as one joint state and solved, the network is restored between pairs, and every
named pair is evaluated. An answer read off a single-outage screen fails even when its number is
close, because post-outage flows do not add up across outages.
\exfield{Result}
Every tool-using system answers all seven RL2Grid tasks. Of the frontier base models,
GPT-6.1 sol answers two and Fable 5.1 one, and the work of neither is verified on any.
\end{example}

\section{Power-System Metrics}
\label{app:metrics}

Each power-system task is graded on the answer a system reports and on the work that produced it.
This appendix defines the terms of Eq.~\ref{eq:e2e}, how the work is verified and why the terms are
weighted as they are.

\paragraph{Why the work is graded.} Most answers are voltages, loadings or costs on standard IEEE
and PGLib networks, so a value can be recalled from published results, guessed close to the truth or
reached by a study on the wrong grid state. An engineer acts on a result only when the study behind it is known
to be right, so only a correct answer with verified work earns full credit.

\paragraph{Terms for one task.} For task $i$, let $a_i$ be the reported value and $g_i$ the answer
key, compared at the benchmark's own tolerance.
\begin{itemize}[leftmargin=*,itemsep=1pt]
  \item \emph{Correct answer}: $r_i=1$ if $a_i$ is within tolerance of $g_i$ and every named choice,
  such as the outage pair in Example~\ref{ex:rl2grid}, matches the key, and $r_i=0$ otherwise.
  \item \emph{Verified work}: $v_i=1$ if the recorded work shows the required study on the required
  grid state, and $v_i=0$ otherwise.
  \item \emph{Strict pass}: $s_i=r_i\,v_i$.
  \item \emph{Numeric closeness}: $c_i=1$ within tolerance, and
  $\max\bigl(0,\,1-|a_i-g_i|/\max(|g_i|,1)\bigr)$ otherwise.
\end{itemize}
$A$, $P$ and $\bar c$ are the means of $r_i$, $s_i$ and $c_i$ over all $n$ tasks, so $E$ is the mean
of $e_i=0.25\,c_i+0.25\,r_i+0.5\,s_i$. A task scores 1 when its answer is correct and its work
verified, 0.5 when only its answer is correct, at most 0.25 when its answer is wrong and 0 when it
has no answer. Tasks a system cannot execute, for example for lack of a tool, stay in the mean.
Leaving them out would grade that system on an easier subset.

\paragraph{Verifying the work.} The required work comes from the task, never from the answer key:
the grid state the question fixes (the published network with every stated change), the type of
study, any stated study arguments and, for a comparison, every candidate state.
Table~\ref{tab:verify} shows what is checked for each kind of system. A value the base case already
holds may be restated or left as it is, but any other change, such as an extra open line, makes a
different state and fails.

\begin{table}[h]
\centering
\caption{\textbf{How the work of each kind of system is verified.}}
\label{tab:verify}
\footnotesize
\setlength{\tabcolsep}{4pt}
\begin{tabular}{@{}>{\raggedright\arraybackslash}p{1.7cm} >{\raggedright\arraybackslash}p{2.5cm} >{\raggedright\arraybackslash}p{3.3cm}@{}}
\toprule
System & Record checked & Work is verified when \\
\midrule
Grid Studio & Its tool calls, replayed & A study of the required type ran on exactly the required state \\ \addlinespace[3pt]
Power-Agent & Every state its solver returned, logged in the sandbox & A logged state matches our own solution of each required network \\ \addlinespace[3pt]
Claude Code, Codex CLI & The solved states they report with their answer & Each required state matches our own solution of the required network \\ \addlinespace[3pt]
Base models & The solved states they report with their answer & As for Claude Code and Codex CLI \\
\bottomrule
\end{tabular}
\end{table}

The coding agents and the base models solve the network in their own code or by
themselves, so they report with their answer the solved states it rests on. We solve the required
network ourselves and accept a power flow when the reported voltages match ours at every bus within
$10^{-4}$\,pu and $0.01^\circ$, up to a common angle shift, an N-1 or a comparison when each of its
states does, and an OPF when its dispatch balances and respects every limit of the required network.
In a control on 31 sampled tasks, this check accepted every correct state and rejected states solved
with a required change left out, apart from two OPF tasks where that change does not bind. Base models
receive the network as text in their prompt, or the question alone where the network
is too large to fit.

\paragraph{Why these weights.} Strict pass carries half the weight because it is the result that
matters in use, so a system whose work cannot be shown reaches at most half the score. Answer
accuracy carries a quarter, so a correct answer without verified work still beats a wrong one.
Closeness carries the last quarter. It equals 1 on every correct answer and only separates a near
miss from a large error. A guess at a per-unit voltage near 1 earns much of it, so it never decides
a pass.

\section{Full Results on the Power-System Benchmarks}
\label{app:accuracy}

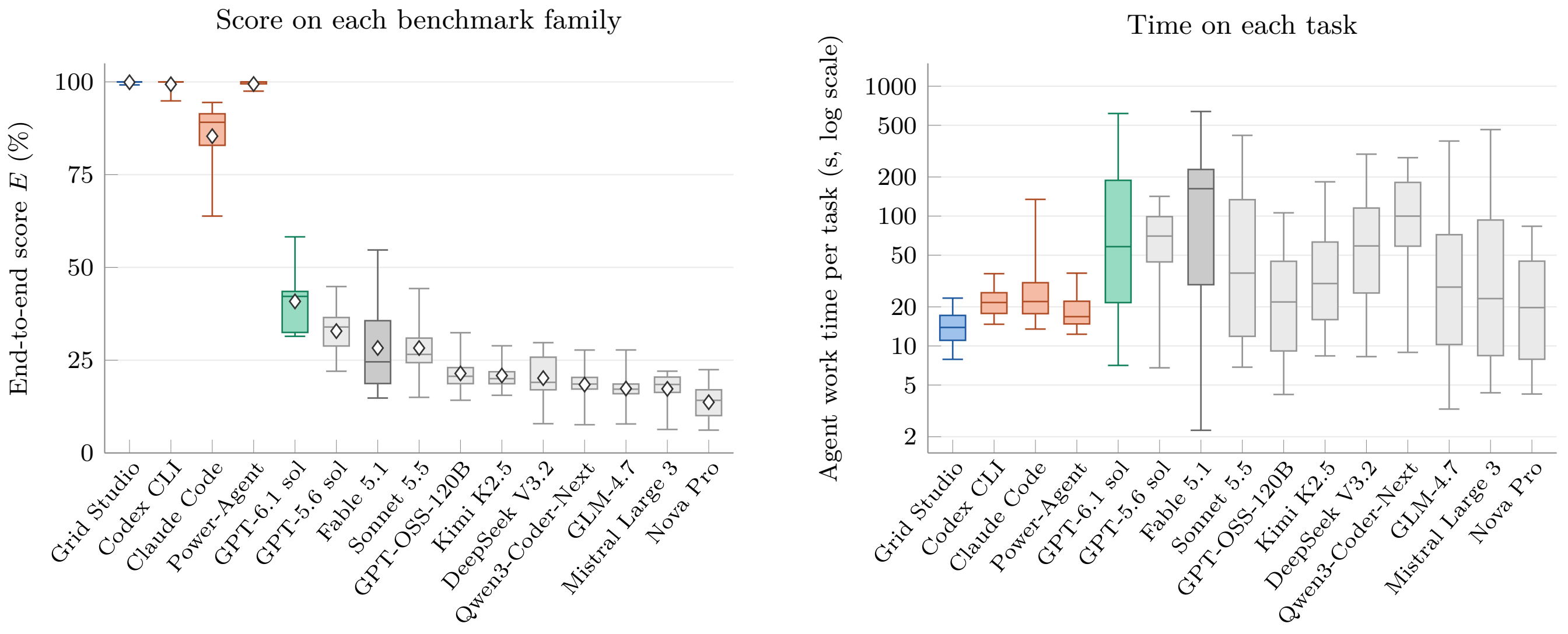
\begin{figure*}[t]
\centering
\begin{tikzpicture}
\begin{axis}[gaplot, width=0.49\textwidth, height=6.0cm, xmin=0.4, xmax=15.6, xtick={1,...,15}, xticklabels={{Grid Studio},{Codex CLI},{Claude Code},{Power-Agent},{GPT-6.1 sol},{GPT-5.6 sol},{Fable 5.1},{Sonnet 5.5},{GPT-OSS-120B},{Kimi K2.5},{DeepSeek V3.2},{Qwen3-Coder-Next},{GLM-4.7},{Mistral Large 3},{Nova Pro}}, x tick label style={rotate=50, anchor=north east, inner sep=1pt, font=\scriptsize}, ymajorgrids, boxplot/draw direction=y, ymin=0, ymax=105, ytick={0,25,50,75,100},
  ylabel={End-to-end score $E$ (\%)}, title={\small Score on each benchmark family}]
\addplot+[boxplot prepared={draw position=1, lower whisker=99.190, lower quartile=100.000, median=100.000, upper quartile=100.000, upper whisker=100.000, box extend=0.62}, solid, fill=vzours!45, draw=vzours!75!black, line width=0.5pt, mark=none] coordinates {};
\addplot[only marks, mark=diamond*, mark size=2.3pt, fill=white, draw=black!80, line width=0.5pt] coordinates {(1,99.899)};
\addplot+[boxplot prepared={draw position=2, lower whisker=94.877, lower quartile=100.000, median=100.000, upper quartile=100.000, upper whisker=100.000, box extend=0.62}, solid, fill=vzcode!45, draw=vzcode!75!black, line width=0.5pt, mark=none] coordinates {};
\addplot[only marks, mark=diamond*, mark size=2.3pt, fill=white, draw=black!80, line width=0.5pt] coordinates {(2,99.360)};
\addplot+[boxplot prepared={draw position=3, lower whisker=63.832, lower quartile=82.907, median=89.113, upper quartile=91.405, upper whisker=94.453, box extend=0.62}, solid, fill=vzcode!45, draw=vzcode!75!black, line width=0.5pt, mark=none] coordinates {};
\addplot[only marks, mark=diamond*, mark size=2.3pt, fill=white, draw=black!80, line width=0.5pt] coordinates {(3,85.370)};
\addplot+[boxplot prepared={draw position=4, lower whisker=97.510, lower quartile=99.472, median=100.000, upper quartile=100.000, upper whisker=100.000, box extend=0.62}, solid, fill=vzcode!45, draw=vzcode!75!black, line width=0.5pt, mark=none] coordinates {};
\addplot[only marks, mark=diamond*, mark size=2.3pt, fill=white, draw=black!80, line width=0.5pt] coordinates {(4,99.425)};
\addplot+[boxplot prepared={draw position=5, lower whisker=31.432, lower quartile=32.476, median=42.192, upper quartile=43.526, upper whisker=58.239, box extend=0.62}, solid, fill=vzgpt!45, draw=vzgpt!75!black, line width=0.5pt, mark=none] coordinates {};
\addplot[only marks, mark=diamond*, mark size=2.3pt, fill=white, draw=black!80, line width=0.5pt] coordinates {(5,40.843)};
\addplot+[boxplot prepared={draw position=6, lower whisker=22.021, lower quartile=28.824, median=33.942, upper quartile=36.505, upper whisker=44.850, box extend=0.62}, solid, fill=vzbase!18, draw=vzbase!90, line width=0.5pt, mark=none] coordinates {};
\addplot[only marks, mark=diamond*, mark size=2.3pt, fill=white, draw=black!80, line width=0.5pt] coordinates {(6,32.835)};
\addplot+[boxplot prepared={draw position=7, lower whisker=14.796, lower quartile=18.706, median=24.554, upper quartile=35.643, upper whisker=54.701, box extend=0.62}, solid, fill=vzbase!45, draw=vzbase!75!black, line width=0.5pt, mark=none] coordinates {};
\addplot[only marks, mark=diamond*, mark size=2.3pt, fill=white, draw=black!80, line width=0.5pt] coordinates {(7,28.287)};
\addplot+[boxplot prepared={draw position=8, lower whisker=14.989, lower quartile=24.335, median=26.561, upper quartile=30.952, upper whisker=44.303, box extend=0.62}, solid, fill=vzbase!18, draw=vzbase!90, line width=0.5pt, mark=none] coordinates {};
\addplot[only marks, mark=diamond*, mark size=2.3pt, fill=white, draw=black!80, line width=0.5pt] coordinates {(8,28.272)};
\addplot+[boxplot prepared={draw position=9, lower whisker=14.205, lower quartile=18.675, median=20.662, upper quartile=23.002, upper whisker=32.404, box extend=0.62}, solid, fill=vzbase!18, draw=vzbase!90, line width=0.5pt, mark=none] coordinates {};
\addplot[only marks, mark=diamond*, mark size=2.3pt, fill=white, draw=black!80, line width=0.5pt] coordinates {(9,21.424)};
\addplot+[boxplot prepared={draw position=10, lower whisker=15.537, lower quartile=18.657, median=20.024, upper quartile=21.884, upper whisker=28.879, box extend=0.62}, solid, fill=vzbase!18, draw=vzbase!90, line width=0.5pt, mark=none] coordinates {};
\addplot[only marks, mark=diamond*, mark size=2.3pt, fill=white, draw=black!80, line width=0.5pt] coordinates {(10,20.861)};
\addplot+[boxplot prepared={draw position=11, lower whisker=7.887, lower quartile=17.020, median=19.022, upper quartile=25.806, upper whisker=29.722, box extend=0.62}, solid, fill=vzbase!18, draw=vzbase!90, line width=0.5pt, mark=none] coordinates {};
\addplot[only marks, mark=diamond*, mark size=2.3pt, fill=white, draw=black!80, line width=0.5pt] coordinates {(11,20.161)};
\addplot+[boxplot prepared={draw position=12, lower whisker=7.607, lower quartile=17.219, median=18.572, upper quartile=20.361, upper whisker=27.737, box extend=0.62}, solid, fill=vzbase!18, draw=vzbase!90, line width=0.5pt, mark=none] coordinates {};
\addplot[only marks, mark=diamond*, mark size=2.3pt, fill=white, draw=black!80, line width=0.5pt] coordinates {(12,18.427)};
\addplot+[boxplot prepared={draw position=13, lower whisker=7.814, lower quartile=15.963, median=17.191, upper quartile=18.571, upper whisker=27.753, box extend=0.62}, solid, fill=vzbase!18, draw=vzbase!90, line width=0.5pt, mark=none] coordinates {};
\addplot[only marks, mark=diamond*, mark size=2.3pt, fill=white, draw=black!80, line width=0.5pt] coordinates {(13,17.354)};
\addplot+[boxplot prepared={draw position=14, lower whisker=6.335, lower quartile=16.313, median=18.459, upper quartile=20.446, upper whisker=22.038, box extend=0.62}, solid, fill=vzbase!18, draw=vzbase!90, line width=0.5pt, mark=none] coordinates {};
\addplot[only marks, mark=diamond*, mark size=2.3pt, fill=white, draw=black!80, line width=0.5pt] coordinates {(14,17.264)};
\addplot+[boxplot prepared={draw position=15, lower whisker=6.148, lower quartile=10.071, median=14.184, upper quartile=17.022, upper whisker=22.444, box extend=0.62}, solid, fill=vzbase!18, draw=vzbase!90, line width=0.5pt, mark=none] coordinates {};
\addplot[only marks, mark=diamond*, mark size=2.3pt, fill=white, draw=black!80, line width=0.5pt] coordinates {(15,13.672)};
\end{axis}
\begin{axis}[gaplot, width=0.49\textwidth, height=6.0cm, xmin=0.4, xmax=15.6, xtick={1,...,15}, xticklabels={{Grid Studio},{Codex CLI},{Claude Code},{Power-Agent},{GPT-6.1 sol},{GPT-5.6 sol},{Fable 5.1},{Sonnet 5.5},{GPT-OSS-120B},{Kimi K2.5},{DeepSeek V3.2},{Qwen3-Coder-Next},{GLM-4.7},{Mistral Large 3},{Nova Pro}}, x tick label style={rotate=50, anchor=north east, inner sep=1pt, font=\scriptsize}, ymajorgrids, boxplot/draw direction=y, xshift=0.525\textwidth, ymode=log, ymin=1.5, ymax=1500,
  ytick={2,5,10,20,50,100,200,500,1000}, yticklabels={2,5,10,20,50,100,200,500,1000},
  ylabel={Agent work time per task (s, log scale)}, title={\small Time on each task}]
\addplot+[boxplot prepared={draw position=1, lower whisker=7.888, lower quartile=11.046, median=13.889, upper quartile=17.184, upper whisker=23.356, box extend=0.62}, solid, fill=vzours!45, draw=vzours!75!black, line width=0.5pt, mark=none] coordinates {};
\addplot+[boxplot prepared={draw position=2, lower whisker=14.690, lower quartile=17.800, median=21.600, upper quartile=25.700, upper whisker=36.020, box extend=0.62}, solid, fill=vzcode!45, draw=vzcode!75!black, line width=0.5pt, mark=none] coordinates {};
\addplot+[boxplot prepared={draw position=3, lower whisker=13.493, lower quartile=17.723, median=21.976, upper quartile=30.689, upper whisker=134.478, box extend=0.62}, solid, fill=vzcode!45, draw=vzcode!75!black, line width=0.5pt, mark=none] coordinates {};
\addplot+[boxplot prepared={draw position=4, lower whisker=12.31, lower quartile=14.78, median=16.81, upper quartile=22.09, upper whisker=36.35, box extend=0.62}, solid, fill=vzcode!45, draw=vzcode!75!black, line width=0.5pt, mark=none] coordinates {};
\addplot+[boxplot prepared={draw position=5, lower whisker=7.082, lower quartile=21.559, median=58.156, upper quartile=188.562, upper whisker=616.413, box extend=0.62}, solid, fill=vzgpt!45, draw=vzgpt!75!black, line width=0.5pt, mark=none] coordinates {};
\addplot+[boxplot prepared={draw position=6, lower whisker=6.787, lower quartile=44.372, median=70.147, upper quartile=99.045, upper whisker=141.817, box extend=0.62}, solid, fill=vzbase!18, draw=vzbase!90, line width=0.5pt, mark=none] coordinates {};
\addplot+[boxplot prepared={draw position=7, lower whisker=2.241, lower quartile=29.629, median=162.541, upper quartile=228.508, upper whisker=639.006, box extend=0.62}, solid, fill=vzbase!45, draw=vzbase!75!black, line width=0.5pt, mark=none] coordinates {};
\addplot+[boxplot prepared={draw position=8, lower whisker=6.863, lower quartile=11.854, median=36.392, upper quartile=133.764, upper whisker=418.403, box extend=0.62}, solid, fill=vzbase!18, draw=vzbase!90, line width=0.5pt, mark=none] coordinates {};
\addplot+[boxplot prepared={draw position=9, lower whisker=4.231, lower quartile=9.150, median=21.809, upper quartile=44.871, upper whisker=105.916, box extend=0.62}, solid, fill=vzbase!18, draw=vzbase!90, line width=0.5pt, mark=none] coordinates {};
\addplot+[boxplot prepared={draw position=10, lower whisker=8.384, lower quartile=15.952, median=30.240, upper quartile=63.124, upper whisker=183.585, box extend=0.62}, solid, fill=vzbase!18, draw=vzbase!90, line width=0.5pt, mark=none] coordinates {};
\addplot+[boxplot prepared={draw position=11, lower whisker=8.279, lower quartile=25.568, median=58.901, upper quartile=115.321, upper whisker=299.803, box extend=0.62}, solid, fill=vzbase!18, draw=vzbase!90, line width=0.5pt, mark=none] coordinates {};
\addplot+[boxplot prepared={draw position=12, lower whisker=8.923, lower quartile=58.668, median=99.961, upper quartile=181.616, upper whisker=281.727, box extend=0.62}, solid, fill=vzbase!18, draw=vzbase!90, line width=0.5pt, mark=none] coordinates {};
\addplot+[boxplot prepared={draw position=13, lower whisker=3.265, lower quartile=10.254, median=28.393, upper quartile=71.901, upper whisker=378.267, box extend=0.62}, solid, fill=vzbase!18, draw=vzbase!90, line width=0.5pt, mark=none] coordinates {};
\addplot+[boxplot prepared={draw position=14, lower whisker=4.358, lower quartile=8.424, median=23.150, upper quartile=93.291, upper whisker=463.554, box extend=0.62}, solid, fill=vzbase!18, draw=vzbase!90, line width=0.5pt, mark=none] coordinates {};
\addplot+[boxplot prepared={draw position=15, lower whisker=4.264, lower quartile=7.898, median=19.742, upper quartile=44.975, upper whisker=83.500, box extend=0.62}, solid, fill=vzbase!18, draw=vzbase!90, line width=0.5pt, mark=none] coordinates {};
\end{axis}
\end{tikzpicture}
\caption{\textbf{Spread of each system's power-system results.} Left: $E$ on each of the eight benchmark
families (box: middle half, whiskers: all eight, diamond: mean). Right: agent work time on each
of the 519 tasks (box: middle half, line: median, whiskers: 5th to 95th percentile).}
\label{fig:studio-boxes}
\end{figure*}

Table~\ref{tab:studio-full} gives the end-to-end score of every system on the 519 tasks, and
Figure~\ref{fig:studio-boxes} shows how much each system's results vary across families and tasks.

\begin{table*}[t]
\centering
\small
\setlength{\tabcolsep}{3.6pt}
\caption{\textbf{End-to-end score of every system on each benchmark family.} $E$ (\%, Eq.~\ref{eq:e2e}) on
the 519 tasks, with PowerCodeBench (PCB) at 352. $\pm$ is the half-width of Grid Studio's 95\%
confidence interval. Base models get the network as text.}
\label{tab:studio-full}
\begin{tabular}{@{}l c c c c c c c c c@{}}
\toprule
& \multicolumn{9}{c@{}}{End-to-end score $E$ (\%)\,$\uparrow$} \\
\cmidrule(l){2-10}
System & PFBench & \pfdelta & PAB-SS & PSAB & PCB & PGLib & ProOPF & RL2Grid & Mean \\
\midrule
Grid Studio & 100.0\pmci{0.0} & 100.0\pmci{0.0} & 100.0\pmci{0.0} & 100.0\pmci{0.0} & 99.2\pmci{0.9} & 100.0\pmci{0.0} & 100.0\pmci{0.0} & 100.0\pmci{0.0} & \textbf{99.9}\pmci{0.1} \\
\midrule
\multicolumn{10}{@{}l}{\textit{Coding agents and Power-Agent}} \\
Codex CLI & 100.0 & 100.0 & 100.0 & 100.0 & 94.9 & 100.0 & 100.0 & 100.0 & 99.4 \\
Claude Code & 94.5 & 92.5 & 91.0 & 84.4 & 78.5 & 88.3 & 63.8 & 89.9 & 85.4 \\
Power-Agent & 100.0 & 97.9 & 100.0 & 100.0 & 97.5 & 100.0 & 100.0 & 100.0 & 99.4 \\
\midrule
\multicolumn{10}{@{}l}{\textit{Proprietary base models}} \\
GPT-6.1 sol & 45.5 & 32.8 & 41.8 & 58.2 & 42.9 & 42.5 & 31.6 & 31.4 & 40.8 \\
GPT-5.6 sol & 36.5 & 24.6 & 36.6 & 44.8 & 35.0 & 32.9 & 30.2 & 22.0 & 32.8 \\
Fable 5.1 & 24.4 & 16.8 & 36.1 & 54.7 & 14.8 & 19.3 & 35.5 & 24.7 & 28.3 \\
Sonnet 5.5 & 29.3 & 15.0 & 35.9 & 44.3 & 24.2 & 28.5 & 24.6 & 24.4 & 28.3 \\
Nova Pro & 14.1 & 11.3 & 6.1 & 16.7 & 14.3 & 22.4 & 6.4 & 18.0 & 13.7 \\
\midrule
\multicolumn{10}{@{}l}{\textit{Open-weight base models}} \\
GPT-OSS-120B & 17.5 & 14.2 & 24.4 & 32.4 & 20.0 & 22.6 & 21.3 & 19.1 & 21.4 \\
Kimi K2.5 & 21.2 & 15.5 & 24.1 & 28.9 & 18.7 & 21.0 & 19.1 & 18.5 & 20.9 \\
DeepSeek V3.2 & 17.7 & 15.0 & 27.8 & 29.7 & 20.3 & 17.8 & 7.9 & 25.1 & 20.2 \\
Qwen3-Coder-Next & 18.7 & 14.9 & 27.7 & 22.3 & 18.0 & 18.4 & 7.6 & 19.7 & 18.4 \\
GLM-4.7 & 16.1 & 15.6 & 17.9 & 27.8 & 18.6 & 18.6 & 7.8 & 16.5 & 17.4 \\
Mistral Large 3 & 16.8 & 14.8 & 6.3 & 22.0 & 17.4 & 20.3 & 19.6 & 20.9 & 17.3 \\
\bottomrule
\end{tabular}
\end{table*}

\section{Evaluation Protocol Details}
\label{app:protocol}

\paragraph{GB Regulation Set.} Pages and figures are sampled from the corpus with a seeded
generator, never through the system under test. A model sees only the sampled passage or figure and
writes a question, its answer and verbatim quotations. Separate checks discard questions the
evidence does not support and questions answerable without it~\citep{golchin2024timetravel}, and
the domain owner approves the rest. Each question keeps its evidence anchors (document, page,
character span or figure). The 45 questions (25 text, 14 figure, 6 multi-hop) come from 23 of the
27 documents.

\paragraph{ViDoRe V3 depth.} From the benchmark's own labels, the 203 multi-hop questions are R3,
the 1,312 others answered from a table, chart, infographic or image are R2, and the remaining 904
are R1.

\paragraph{Power-system questions.} Every task is posed as a question about a stated network with
one value to report, so that every system receives the same wording and is graded the same way.
PFBench, PowerCodeBench and ProOPF-B publish such questions, and we ask them in their authors'
words. Where one omits a quantity or choice its answer depends on (156 of 352 PowerCodeBench and 6
of 19 ProOPF-B items), we restore it from the item's own record, never adding the answer. \pfdelta{}
and PGLib-OPF publish networks and solved results but no questions, so we pose each task from a
fixed template that names the network, any stated changes and the quantity to report.
PowerAgentBench-SS releases task specifications without a question set, PSAB's tasks are built for
agents that work on files, and RL2Grid scores reinforcement-learning episodes rather than answers.
For these three we keep the kind of study each one tests and write the questions ourselves, with
answers from a reference solver. In all, 389 of the 519 questions are in their authors' words and
130 are ours. Ours come from fixed templates, not from a language model.

\paragraph{Answer keys.} Answer keys never come from the system under test. The regulatory keys are
the published ones, apart from the GB Regulation Set's, which the domain owner approved. PFBench,
PowerCodeBench and ProOPF-B use their authors' published answers, \pfdelta{} its authors' Ipopt
solutions, and PGLib-OPF its published OPF baseline, with a reference solver for its power-flow
tasks. The three adapted families (Table~\ref{tab:benchmarks}) are graded against a reference
solver, so we treat them as weaker evidence. Run with no model, the reference solver reproduces
every published answer of PFBench, \pfdelta{}, PowerCodeBench and ProOPF-B (460 tasks), so a solver
discrepancy is never counted as an agent failure.

\paragraph{Isolation.} Each coding-agent task runs in its own container (2 vCPU, 8\,GB) with no
answer key and no network access, and a task that fetched anything would count as unanswered. None
did. An audit of every GB Regulation Set run with file access found no read of an answer key. Grid
Studio runs with memory off, and its timed runs use the same container size, one task per
container.

\paragraph{Agent work time.} Agent work is the time spent on model turns and tool runs, read from
each agent's own clock. It excludes
start-up, which covers launching and tearing down the container and starting the agent and, for
Grid Studio, its solvers. Codex CLI reports no such split, so its time is the whole run of
its process, an upper bound. A base model spends all its time on one completion.
Figure~\ref{fig:frontier} reports the median over the 519 tasks. A coding agent loads its libraries
inside its own scripts, while starting Grid Studio's solvers, 6.7\,s on average when we measured
it, is not counted.

\paragraph{Systems compared.} Table~\ref{tab:systems} gives each system's version and setting. The
two base models in the table and the nine further ones in Figure~\ref{fig:frontier} and
Table~\ref{tab:studio-full} ran on the same 519 tasks between 29 and 31 October 2026, each with the
network in its prompt, and were graded with the same grader. Grid Studio, Power-Agent and the base models
ran once per task, and Codex CLI and Claude Code once on PowerCodeBench and five times on the other
families. Power-Agent ran on 2 November 2026 in the same sandbox as the coding agents, with no
network access beyond the model. GPT-5.6 sol lost 26 PowerCodeBench tasks and GPT-6.1 sol 3 to provider quota limits,
counted as unanswered, and Fable 5.1 answered 146 PowerCodeBench questions in their earlier
wording. Fable 5.1 and Sonnet 5.5 reached the reply limit without an answer on 170 and 128 tasks,
and Fable 5.1 refused 115. Where a provider's quota ran out during a run,
the remaining tasks ran on the same model through another provider (GPT-6.1 sol and Sonnet 5.5).
Grid Studio's tools were extended where a benchmark, PowerCodeBench among them, asked for a study
they did not yet cover, and the affected benchmarks were run again.

\begin{table}[h]
\centering
\small
\caption{\textbf{Systems compared on the power-system benchmarks.}}
\label{tab:systems}
\footnotesize
\setlength{\tabcolsep}{4pt}
\begin{tabular}{@{}l >{\raggedright\arraybackslash}p{5.6cm}@{}}
\toprule
System & Configuration \\
\midrule
Grid Studio & With its domain tools, memory off \\
Claude Code & CLI v2.1.289 on the same model as Grid Studio, with shell and Python (NumPy, SciPy) \\
Codex CLI & CLI v0.160.0 on GPT-6.1 sol, with shell and Python (NumPy, SciPy) \\
Power-Agent & Claude Code CLI v2.1.289 on the same model as Grid Studio, with PowerMCP~0.4.1, the PowerSkills playbooks as published, shell, Python and pandapower \\
GPT-6.1 sol & One completion, no tools, network as text in the prompt \\
Fable 5.1 & Claude Fable 5.1, one completion, no tools, network as text in the prompt \\
\bottomrule
\end{tabular}
\end{table}

\section{LegalBench-RAG Details}
\label{app:legalbench}

Table~\ref{tab:ontology} scores LegalBench-RAG by document. Precision is the share of cited
contracts that hold the answer, and recall the share of answer-holding contracts that were cited.
The benchmark's own metrics are stricter and score the character overlap between the cited text and
the gold clause. Table~\ref{tab:legalbench-full} gives both.

At the clause level, agentic systems reach an F1 an order of magnitude above retrieval pipelines
(0.47 to 0.62 against 0.03 to 0.09), so pipelines find the right contract but not the clause. In a
sweep over chunking, hybrid search and embeddings~\citep{xiao2024cpack}, no pipeline
configuration exceeded a clause F1 of 0.13. PageIndex reaches clause recall 0.72 but precision 0.05, because it returns
whole sections. The Grid Ontology leads on both clause and document F1, and Example~\ref{ex:cuad}
shows what clause-level scoring rewards.

\begin{table}[h]
\centering
\small
\caption{\textbf{LegalBench-RAG at the clause level.} \emph{Span} metrics are the benchmark's
own, scored on characters of the gold clause, and \emph{Doc F1} is the F1 of
Table~\ref{tab:ontology}. All systems answer the same 50 questions.}
\label{tab:legalbench-full}
\footnotesize
\setlength{\tabcolsep}{2pt}
\begin{tabular}{@{}l c c c c@{}}
\toprule
& \multicolumn{3}{c}{Span} & Doc \\
\cmidrule(lr){2-4}\cmidrule(l){5-5}
System & P\,$\uparrow$ & R\,$\uparrow$ & F1\,$\uparrow$ & F1\,$\uparrow$ \\
\midrule
Grid Ontology & 0.587 & 0.748 & \textbf{0.616} & \textbf{0.980} \\
Claude Code (file tools) & 0.570 & 0.679 & 0.561 & 0.900 \\
\midrule
Recursive LM & -- & -- & 0.467 & 0.860 \\
Vector + reranker & 0.029 & 0.343 & 0.050 & 0.869 \\
PageIndex, official & 0.050 & 0.723 & 0.087 & 0.799 \\
Dense hybrid, semantic & -- & -- & 0.047 & 0.704 \\
Dense hybrid, parent-child & -- & -- & 0.034 & 0.796 \\
\bottomrule
\end{tabular}
\end{table}

\section{ViDoRe V3 Details}
\label{app:vidore}

\begin{table}[t]
\centering
\small
\caption{\textbf{ViDoRe V3 nDCG@10 ($\times 100$) per set.} Ours is the Grid Ontology. VAR is
VisAgentRet (EVIE-4.5B with a GLM-5.3-Flash agent), NeMo is NeMo Retriever's agentic pipeline
(ColEmbed-VL-8B with a Claude Opus 4.5 agent), JEV is EVIE-4.5B reranked by JEV-27B-VL, and Jina is
jina-embeddings-v4 reranked by zerank-2. Best in \textbf{bold}.}
\label{tab:vidore-sets}
\footnotesize
\setlength{\tabcolsep}{3pt}
\begin{tabular}{@{}l r r r r r@{}}
\toprule
& \multicolumn{5}{c@{}}{nDCG@10 ($\times$100)\,$\uparrow$} \\
\cmidrule(l){2-6}
Set & Ours & VAR & NeMo & JEV & Jina \\
\midrule
Finance (EN) & \textbf{77.62} & 76.67 & 75.88 & 74.49 & 74.24 \\
Finance (FR) & \textbf{60.94} & 60.30 & 58.60 & 56.96 & 51.43 \\
Computer science & \textbf{85.60} & 84.81 & 84.49 & 84.01 & 83.46 \\
Human resources & 73.48 & \textbf{74.51} & 74.50 & 71.68 & 69.10 \\
Energy & 75.36 & \textbf{75.68} & 75.52 & 73.42 & 71.17 \\
Industrial & \textbf{65.49} & 63.95 & 63.40 & 63.57 & 58.89 \\
Pharmaceuticals & 74.67 & \textbf{75.14} & 74.79 & 72.42 & 68.58 \\
Physics & 48.66 & 49.01 & 46.59 & \textbf{49.42} & 48.06 \\
\midrule
Mean & \textbf{70.23} & 70.01 & 69.22 & 68.25 & 65.62 \\
\bottomrule
\end{tabular}
\end{table}

Table~\ref{tab:vidore-sets} gives the ViDoRe V3 scores of Table~\ref{tab:ontology} per set. The
Grid Ontology leads on four of the eight sets, VisAgentRet on three and EVIE-4.5B with JEV-27B-VL
on physics. We chose the agent's settings on the first 40 questions of each set, then ran all 2,419
questions. One question ended in an error and is left out of its set's mean (counted as zero, the
mean is 70.20).

\end{document}